\documentclass[10pt,twocolumn,letterpaper]{article}

\usepackage{iccv}
\usepackage{times}
\usepackage{epsfig}
\usepackage{graphicx}
\usepackage{amsmath}
\usepackage{amssymb}
\usepackage{booktabs}
\usepackage{bbding}
\usepackage{caption}
\usepackage{subcaption}
\usepackage{array}
\usepackage{multirow}
\usepackage{subfloat}
\usepackage{float}
\usepackage{color}
\usepackage{comment}
\usepackage{colortbl}
\usepackage{pifont} 

\def\x{$\times$}

\newcolumntype{x}[1]{>{\centering\arraybackslash}p{#1pt}}
\newcolumntype{y}[1]{>{\raggedright\arraybackslash}p{#1pt}}
\newcolumntype{z}[1]{>{\raggedleft\arraybackslash}p{#1pt}}
\newcolumntype{k}[1]{>{\raggedright\arraybackslash}p{#1pt}}

\newlength\savewidth\newcommand\shline{\noalign{\global\savewidth\arrayrulewidth\global\arrayrulewidth 1pt}\hline\noalign{\global\arrayrulewidth\savewidth}}

\newcommand{\blockatt}[3]{\multirow{2}{*}{\(\left[\begin{array}{c}\text{MHSA(\wcolor{#1})}\\[-.1em] \text{FFN(\wcolor{#2})}\end{array}\right]\)$\times$#3}
}

\newcommand{\blocktoken}[3]{\multirow{3}{*}{\(\left[\begin{array}{c}\text{MHSA(\wcolor{#1})}\\[-.1em] \text{keep rate=\maskcolor{$\rho$}}\\[-.1em] \text{FFN(\wcolor{#2})}\end{array}\right]\)$\times$#3}
}

\definecolor{xycolor}{RGB}{60, 120, 216}
\newcommand{\xycolor}[1]{\textcolor{xycolor}{#1}}
\definecolor{wcolor}{RGB}{103, 78, 167}
\newcommand{\wcolor}[1]{\textcolor{wcolor}{#1}}
\definecolor{dcolor}{RGB}{166, 77,21}

\definecolor{gcolor}{RGB}{204, 102, 153}

\definecolor{tcolor}{RGB}{34,139,34}
\newcommand{\tcolor}[1]{\textcolor{tcolor}{#1}}
\definecolor{iterc}{RGB}{91,196,159}

\definecolor{epochc}{RGB}{96,172,252}

\definecolor{eicolor}{RGB}{153, 51, 102}

\definecolor{orange}{RGB}{237, 125, 49}
\newcommand{\maskcolor}[1]{\textcolor{orange}{#1}}

\usepackage[pagebackref=false,breaklinks=true,colorlinks,urlcolor=blue,citecolor=blue,linkcolor=blue,bookmarks=false]{hyperref}

\definecolor{mygray}{gray}{0.92}
\definecolor{baselinecolor}{gray}{.9}

\usepackage[capitalize]{cleveref}
\crefname{section}{Sec.}{Secs.}
\Crefname{section}{Section}{Sections}
\Crefname{table}{Table}{Tables}
\crefname{table}{Tab.}{Tabs.}
\usepackage[accsupp]{axessibility}

\iccvfinalcopy % *** Uncomment this line for the final submission

\def\iccvPaperID{5968} % *** Enter the ICCV Paper ID here

\ificcvfinal\fi

\begin{document}

%%%%%%%%% TITLE
\title{Efficient Video Action Detection with Token Dropout and Context Refinement}

\author{Lei Chen$^{1}$ \quad
Zhan Tong$^{2}$ \quad
Yibing Song$^{3}$ \quad
Gangshan Wu$^{1}$ \quad
Limin Wang$^{1,4}$\thanks{Corresponding author.}\\
$^{1}$State Key Laboratory for Novel Software Technology, Nanjing University \\
$^{2}$Ant Group \quad 
$^{3}$AI$^3$ Institute, Fudan University \quad
$^{4}$Shanghai AI Lab \\
\tt\small{leichen1997@outlook.com \quad zhantong.2023@gmail.com
 \quad yibingsong.cv@gmail.com}\vspace{-0.4em} \\ \tt\small{gswu@nju.edu.cn \quad lmwang@nju.edu.cn}
}

\maketitle
\ificcvfinal\thispagestyle{empty}\fi

%%%%%%%%% ABSTRACT
\begin{abstract}
Streaming video clips with large-scale video tokens impede vision transformers (ViTs) for efficient recognition, especially in video action detection where sufficient spatiotemporal representations are required for precise actor identification. In this work, we propose an end-to-end framework for efficient video action detection (EVAD) based on vanilla ViTs. Our EVAD consists of two specialized designs for video action detection. First, we propose a spatiotemporal token dropout from a keyframe-centric perspective. In a video clip, we maintain all tokens from its keyframe, preserve tokens relevant to actor motions from other frames, and drop out the remaining tokens in this clip. Second, we refine scene context by leveraging remaining tokens for better recognizing actor identities. The region of interest (RoI) in our action detector is expanded into temporal domain. The captured spatiotemporal actor identity representations are refined via scene context in a decoder with the attention mechanism. These two designs make our EVAD efficient while maintaining accuracy, which is validated on three benchmark datasets (i.e., AVA, UCF101-24, JHMDB). Compared to the vanilla ViT backbone, our EVAD reduces the overall GFLOPs by 43\% and improves real-time inference speed by 40\% with no performance degradation. Moreover, even at similar computational costs, our EVAD can improve the performance by 1.1 mAP with higher resolution inputs. Code is available at \href{https://github.com/MCG-NJU/EVAD}{https://github.com/MCG-NJU/EVAD}.
\end{abstract}

%%%%%%%%% BODY TEXT
\begin{figure}[t]
\centering
\includegraphics[width=1.\linewidth]{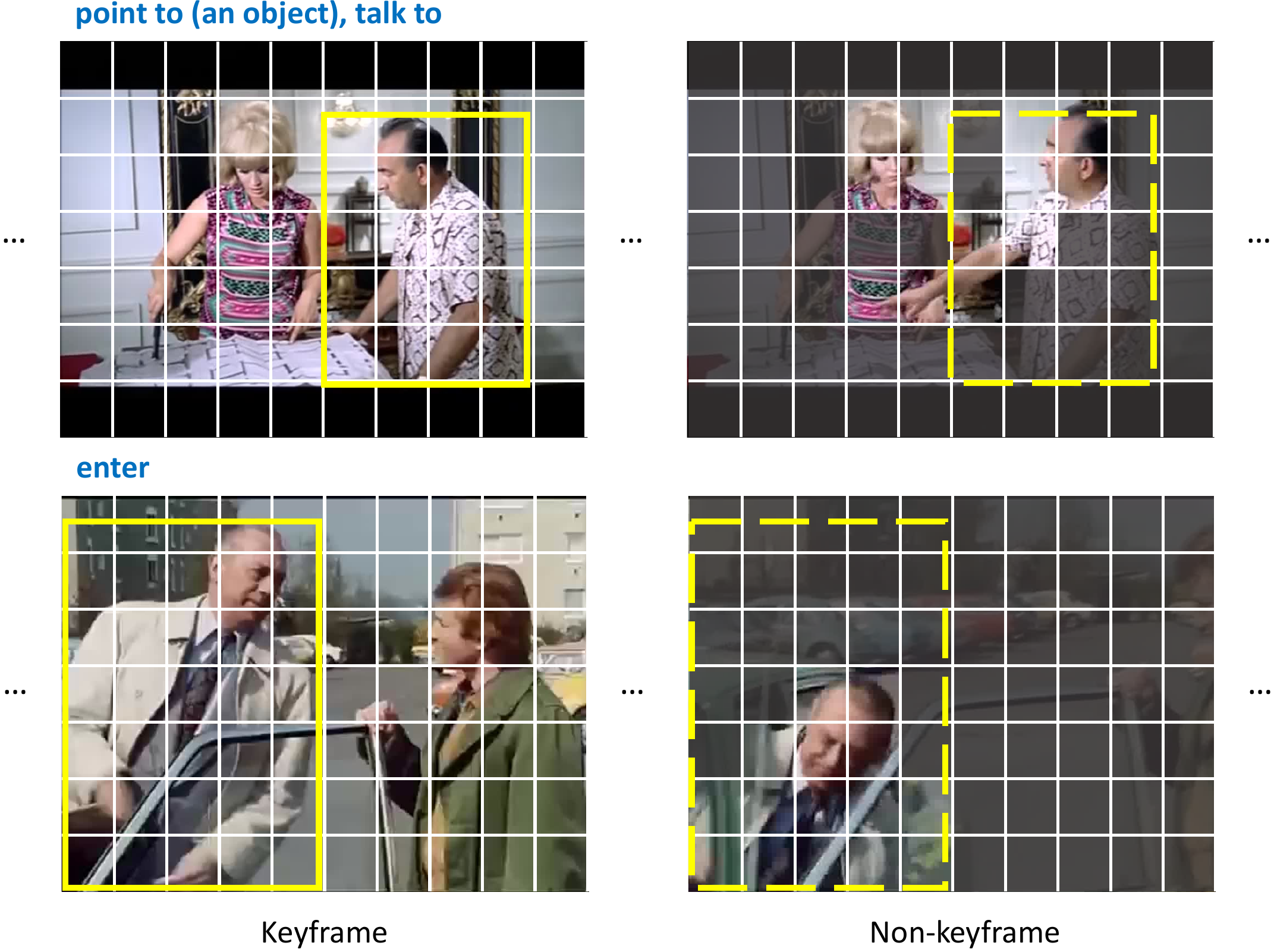}
\vspace{-4mm}
\caption{Spatiotemporal token dropout from a keyframe-centric perspective. We maintain tokens from the keyframe of a video clip, preserve a small amount of tokens from non-keyframes based on actor motions, and drop out the remaining tokens in this clip. On the first row, we preserve tokens relevant to the waving hand in non-keyframes as they benefit recognizing the action `point to'. On the second row, we drop out tokens irrelevant to the action `entering' in non-keyframes for efficient recognition.}
\label{fig:intro}
\vspace{-6mm}
\end{figure}

\section{Introduction}
\label{sec:intro}
Vision transformers (ViTs) have improved a series of recognition performance, including image classification~\cite{dosovitskiy2020image,touvron2021training,jiang2021all} and object detection~\cite{carion2020end,dai2021up,song2021vidt}. The image patches are regarded as tokens for ViT inputs for self-attention computations. When recognizing a video clip, we notice that the tokens are from each frame and thus formulate a large-scale input for ViTs. These video tokens introduce heavy computations during training and inference, especially in the self-attention computation step. While attempts have been made to reduce vision tokens~\cite{rao2021dynamicvit,liang2022not,fayyaz2022adaptive,wang2022efficient} for fast computations, it is still challenging for video action detection (VAD)~\cite{soomro2012ucf101,gu2018ava,li2021multisports} to balance accuracy and efficiency. This is because in VAD we need to localize actors in each frame and recognize their corresponding identities. For each actor, the temporal motion in video sequences shall be maintained for consistent identification. Meanwhile, the scene context ought to be kept to differentiate from other actors. Sufficient video tokens representing both actor motions and scene context will preserve VAD accuracy, which leaves a research direction on how to select them for efficient VAD.

In this paper, we preserve video tokens representing actor motions and scene context, while dropping out irrelevant tokens. Based on the temporal coherency of video clips, we propose a spatiotemporal token dropout from a keyframe-centric perspective. For each video clip, we can select one keyframe representing the scene context where all the tokens shall be maintained. Meanwhile, we select tokens from other frames representing actor motions. Moreover, we drop out the remaining video tokens in this clip. Fig.~\ref{fig:intro} shows two examples. On the first row, we maintain all the tokens in the keyframe,  preserve the tokens in non-keyframe relating to the eyes and mouths associating to the action `talk to', the waving hand of the right person associating to the action `point to', and drop out the remaining video tokens. On the second row, we maintain all the tokens in the keyframe, preserve the tokens in non-keyframes relating to the movement of entering the car from outside which is associated with the action `enter', and drop out the remaining video tokens. Our spatiotemporal token dropout maintains the relevant actor motions and scene context from the coherent video clips, while discarding the remaining irrelevant ones for efficient computations.

We develop spatiotemporal token dropout via a keyframe-centric token pruning module within the ViT encoder backbone. The keyframe is either uniformly sampled or manually predefined in the video clips. We select the middle frame of the input clip with box annotations as the keyframe by default. The feature map of the selected keyframe is retained completely for actor localization. Besides, we extract an attention map enhanced by this keyframe. This attention map guides token dropout in the non-keyframes. After localization, we need to classify each localized bounding box (bbx) for actor identification. As video tokens are incomplete in non-keyframes, the classification performance is inferior in the bbx regions where tokens have been dropped out. Nevertheless, the inherent temporal consistency in video benefits us to refine both actor and scene context from the remaining video tokens. We expand the localized bbxs in the temporal domain for RoIAlign~\cite{DBLP:conf/iccv/HeGDG17} to capture token features related to actor motions. Then, we introduce a decoder to refine actor features guided by the remaining video tokens in this clip. The decoder concatenates the actor and token features, and performs self-attention to produce enriched actor features for better identification. We find that after token dropout, the degraded action classification can be effectively recovered by using remaining video tokens for context refinement. The recovered performance is the same as that using the whole video tokens for action classification. Through this context refinement, we can maintain the VAD performance using reduced video tokens for efficient computations.

We conduct extensive experiments on three popular action detection datasets (AVA~\cite{gu2018ava}, UCF101-24~\cite{soomro2012ucf101}, JHMDB~\cite{jhuang2013towards}) to show the advantages of our proposed EVAD. In the ViT-B encoder backbone, for instance, we employ a keyframe-centric token pruning module three times with a keep rate $\rho$ of 0.7 (i.e,. the dropout rate is 0.3). The encoder outputs 34\% of the original video tokens, which reduces GFLOPs by 43\% and increases throughput by 40\% while achieving on-par performance. On the other hand, under the similar computation cost (i.e., a similar amount of tokens), we can take video clips in a higher resolution for performance improvement. For example, we can improve detection performance by 1.1 mAP when increasing the resolution from 256 to 288 on the short side with GFLOPs reduced by 22\% and throughput increased by 10\%. We also provide visualizations and ablation studies to show how our EVAD performs spatiotemporal token dropout to eliminate action irrelevant video tokens.

\begin{figure*}[t]
\centering
\includegraphics[width=.99\linewidth]{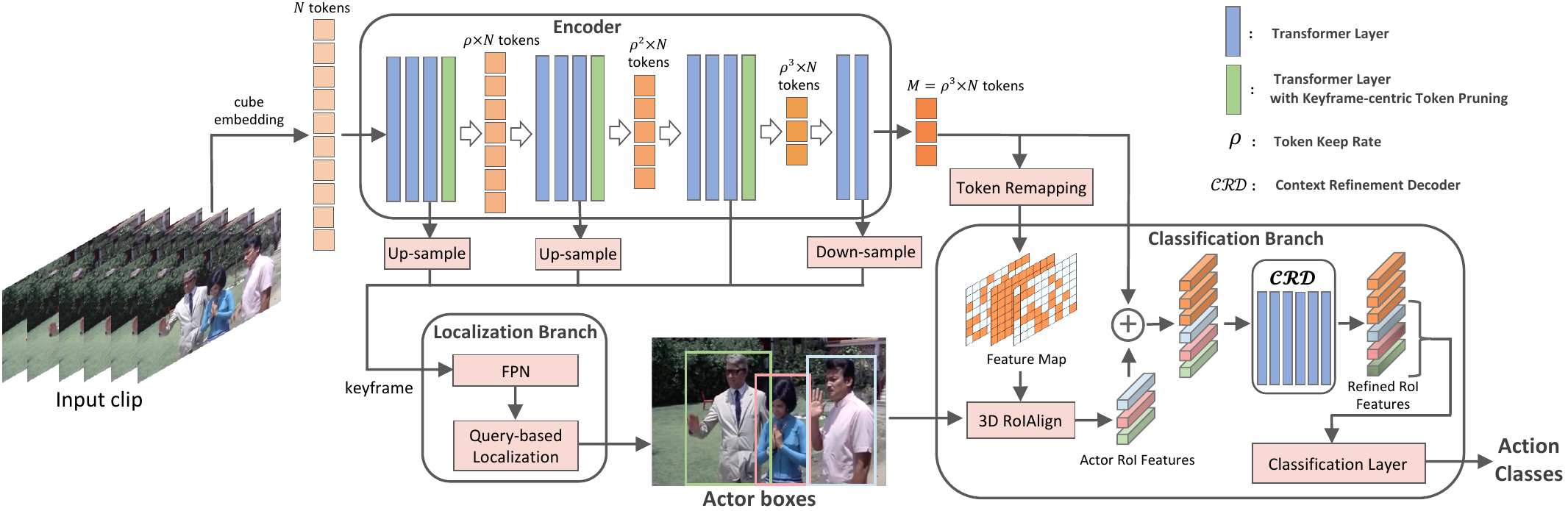}
\vspace{-2mm}
\caption{\textbf{Pipeline of EVAD.} Our detector consists of three parts: an encoder with keyframe-centric token pruning for efficient video feature extraction, a query-based localization branch using multiscale features of the keyframe for actor boxes prediction, and a classification branch conducting actor spatiotemporal feature refinement and relational modeling between actor RoI features and compact context tokens from ViT encoder.}
\label{fig:pipe}
\vspace{-4mm}
\end{figure*}

\section{Related Works}
\subsection{Spatio-temporal Action Detection}
Current state-of-the-art methods~\cite{DBLP:conf/iccv/Feichtenhofer0M19,feichtenhofer2020x3d,DBLP:conf/cvpr/WuF0HKG19,DBLP:conf/eccv/TangXMPL20,pan2021actor,chen2023cycleacr} commonly adopt a two-stage pipeline with two separated backbones, i.e., a 2D backbone for actor localization on keyframes and a 3D backbone for video feature extraction. Some previous approaches~\cite{kopuklu2019you,sun2018actor} simplified the pipeline by training two backbones in an end-to-end manner, which suffer from heavy complexity and optimization difficulty. Most recent methods~\cite{girdhar2019video,chen2021watch,zhao2022tuber,wu2023stmixer} utilize a unified backbone to perform action detection. VAT~\cite{girdhar2019video} is a transformer-style action detector to aggregate the spatiotemporal context around the target actors. WOO~\cite{chen2021watch} and TubeR~\cite{zhao2022tuber} are query-based action detectors following the detection frameworks of~\cite{sun2021sparse,carion2020end} to predict actor bounding boxes and action classes. STMixer~\cite{wu2023stmixer} is a one-stage query-based detector to adaptively sample discriminative features. Several newly transformer-based methods~\cite{fan2021multiscale,wu2022memvit,tong2022videomae} apply a ViT variant backbone and achieve competitive results following the two-stage pipeline. 
Moreover, there are also emerging methods~\cite{arnab2022beyond,kumar2022end,xiao2022hierarchical} that focus on the training paradigm of deep networks for action detection task.

\subsection{Spatio-temporal Redundancy}
{\flushleft \bf Spatial redundancy.} The success of vision transformers has inspired various works~\cite{marin2021token,heo2021rethinking,rao2021dynamicvit,ryoo2021tokenlearner,liang2022not,renggli2022learning,fayyaz2022adaptive} to explore the spatial redundancy of intermediate tokens. 
%Token Pooling~\cite{marin2021token} clusters similar tokens to reduce the number of tokens.
DynamicViT~\cite{rao2021dynamicvit} observes that accurate image recognition is based on a subset of most informative tokens and designs a dynamic token sparsification framework to prune redundant tokens. EViT~\cite{liang2022not} calculates the attentiveness of the class token to each token and identifies the top-k tokens using the attentiveness value. ATS~\cite{fayyaz2022adaptive} introduces a differentiable adaptive token sampler for adaptively sampling significant tokens based on the image content.

{\flushleft \bf Spatio-temporal redundancy.} Due to the high redundancy of video datasets, extensive studies~\cite{ocsampler,DSN,wu2019adaframe,korbar2019scsampler,wang2021adaptive,zhi2021mgsampler,fayyaz20213d} have focused on developing efficient video recognition. 
Some recent approaches~\cite{bertasius2021space,arnab2021vivit,bulat2021space,liu2022video,qing2022mar,park2022k,wang2022efficient,wang2023videomae} present efficient schemes specialized for video transformers. For the reduction of video tokens, MAR~\cite{qing2022mar} devises a hand-crafted masking strategy to discard a proportion of patches. K-centered~\cite{park2022k} proposes a patch-based sampling that outperforms the traditional frame-based sampling. STTS~\cite{wang2022efficient} conducts token selection sequentially in temporal and spatial dimensions within the transformer. VideoMAE V2~\cite{wang2023videomae} designs a dual masking strategy for efficient pre-training.

In contrast to these methods, we are the first to investigate redundancy in video action detection. We consider the correlation of keyframe and adjacent frames to drop out redundant tokens, and implement an efficient transformer-based action detector in an end-to-end manner.

\section{Method}
\label{sec:method}
\subsection{Overall EVAD Architecture}
\label{sec:pipe}
{\flushleft \bf Video transformers overview.}
Video transformers~\cite{arnab2021vivit,tong2022videomae} perform tokenization by dividing the video sequence into $T/2\times H/16\times W/16$ cubes, each of size $2\times 16\times 16$. Then, project each cube to a 3D token using cude embedding. All tokens are added with positional encoding and fed into a sequentially-stacked transformer encoder consisting of a multi-head self-attention (MHSA) and a feed-forward network (FFN). The encoder output is used for classification after global average pooling. Alternatively, a learnable [CLS] token is added before the encoder for final prediction.

{\flushleft \bf Our pipeline.}
The pipeline of end-to-end video action detection proposed in this paper is based on the vanilla ViT, as shown in Fig.~\ref{fig:pipe}. To alleviate the computational bottleneck caused by joint space-time attention, we devise an efficient video action detector (EVAD) with an encoder-decoder architecture with respect to the characteristics of action detection. EVAD enables the encoder with token pruning to remove the redundant tokens, and the decoder to refine actor spatiotemporal features. Following the setting in WOO~\cite{chen2021watch}, we utilize multiple intermediate spatial feature maps of the keyframe in the encoder for actor localization, and the spatiotemporal feature map output from the last encoder layer for action classification.
\textbf{Token Pruning}. We design a keyframe-centric token pruning module to progressively reduce the redundancy of video data and ensure that few and promising tokens are delivered for action localization and classification. The specific token pruning module will be detailed in the next subsection.
\textbf{Localization}. We up-sample or down-sample intermediate ViT feature maps of the keyframe to produce multiscale feature maps and send them to feature pyramid network (FPN)~\cite{lin2017feature} for multiscale fusion. The localization branch predicts $n$ candidate boxes in the keyframe by a query-based method, same as in~\cite{sun2021sparse,chen2021watch}.
\textbf{Classification}. We remap the compact context tokens from ViT encoder into a spatiotemporal feature map with a regular structure. Then, we conduct 3D RoIAlign on the restored feature map using extended prediction boxes from the localization branch to obtain $n$ actor RoI features. Subsequently, we utilize a context refinement decoder ($CRD$) for actor feature refinement and relational modeling between actor RoI features and compact context from the encoder, and the refined RoI features are passed through a classification layer for final prediction.

\begin{figure}[t]
\centering
\includegraphics[width=0.95\linewidth]{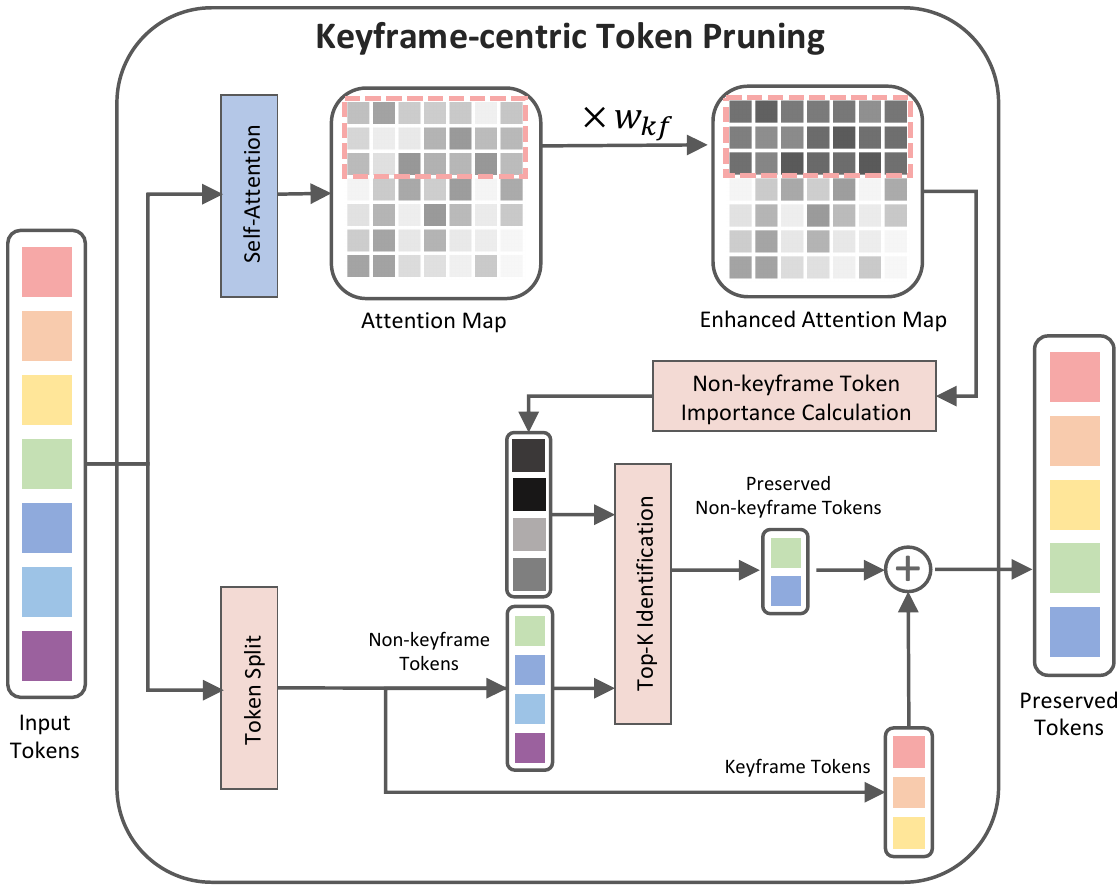}
\vspace{-1mm}
\caption{\textbf{Structure of keyframe-centric token pruning.} We conduct token pruning within the original encoder layer. Specifically, we calculate the importance scores of non-keyframe tokens using a keyframe-enhanced attention map. Then, we preserve the top-k important non-keyframe tokens concatenated with keyframe tokens as the results.}
\label{fig:stru}
\vspace{-3mm}
\end{figure}

\subsection{Keyframe-centric Token Pruning}
\label{sec:token}
The high spatiotemporal redundancy of video with similar semantic information between adjacent frames makes it possible to perform token pruning with a high dropout rate on video transformers. In this paper, we design a keyframe-centric token pruning module, as shown in Fig.~\ref{fig:stru}. We split all spatiotemporal tokens into keyframe tokens and non-keyframe tokens. And we preserve all keyframe tokens for accurate actor localization in the keyframe.

{\flushleft \bf Non-keyframe token pruning.}
For the importance measure of non-keyframe tokens, we refer to the approach of EViT~\cite{liang2022not} in image classification, using the pre-computed attention map to represent the importance of each token without additional learnable parameters and nontrivial computational costs. 
As shown in the top part of Fig.~\ref{fig:stru}, we first average the $num\_heads$ dimension of the attention map to obtain an $N\times N$ matrix, which represents the attentiveness between tokens (omitting the batch size). For example, $attn(i, j)$ denotes that the token i considers the importance of token j.
Since there is no classification token specifically for video-level recognition, we calculate the average importance score of each token by $I_{j}=\frac{1}{N} \sum_{i=1}^{N}attn(i,j)$.
Then we identify the top $(N*\rho-N_{1})$ tokens from the $N_{2}$ non-keyframe tokens in descending order by importance scores, where $N,N_{1},N_{2}$ represent the number of all, keyframe and non-keyframe tokens respectively, and $\rho$ represents token keeping rate.
Normally, the keyframe contains the most accurate semantic information for the current sample, and other frames away from the keyframe incur nontrivial information bias. Thus, it is practical to conduct token pruning guided by the keyframe.
To this end, we insert a {\it Keyframe Attentiveness Enhancement} step between acquiring the attention map and calculating the importance of each non-keyframe token. As presented in Fig.~\ref{fig:stru}, we apply a greater weight value to the keyframe queries, thereby retaining tokens with higher correlations to keyframe tokens. The importance score of each token is updated as follows:
\begin{align}
I_{j}=\frac{1}{N} \sum_{i=1}^{N}\left\{\begin{array}{ll}
{w}_{kf}\cdot\operatorname{attn}(i,j), & i \in\left(0, N_{1}\right) \\
\operatorname{atth}(i,j), & i \in\left(N_{1}, N\right)
\end{array}\right.
\end{align}
where we assume the first $N_{1}$ tokens belong to the keyframe and the weight value ${w}_{kf}$ is a hyper-parameter will be ablated in Sec.~\ref{sec:ablation}.
In other words, we discard some tokens that only have a high response with non-keyframes, which may not be high-quality tokens. Only if the non-keyframe becomes the keyframe of previous/next samples, these highly responsive tokens are high-quality. By dropping out these redundant tokens with a high response to non-keyframes, we can further reduce the number of tokens. After the execution of token pruning, we send the preserved tokens to the follow-up FFN of the encoder layer.

\begin{figure}[t]
\centering
\includegraphics[width=1.\linewidth]{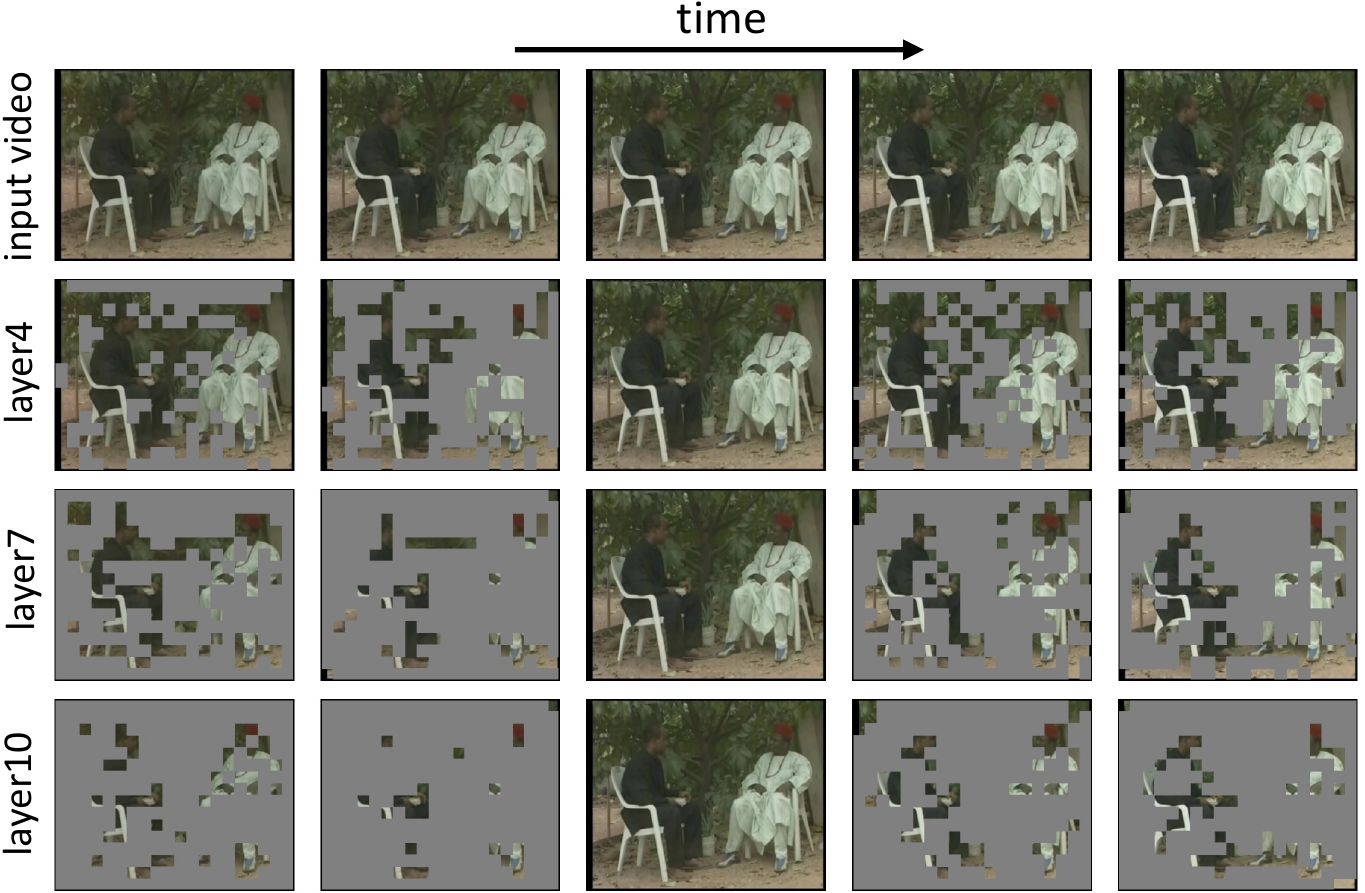}
\caption{\textbf{Visualization of preserved tokens by encoder layers with token pruning.} On the third column, the keyframe tokens are all retained for accurate actor localization. The number of tokens in other frames progressively reduces, whereas important cues such as people and chairs can be effectively retained.}
\label{fig:vis_tp}
\vspace{-3mm}
\end{figure}

The first token pruning is started at 1/3 of encoder layers to ensure that the model is capable of high-level semantic representation. Then, we perform token pruning every 1/4 of the total layers, discarding the redundant tokens and keeping the effective ones. As shown in Fig.~\ref{fig:vis_tp}, we visualize the preserved tokens of each pruning layer in ViT-B, and the model is able to retain important cues such as people and chairs. We present more visualizations in Appendix~\S~D. After multiple pruning, the number of tokens is drastically reduced, which enables the model to save computation costs and accelerate training and inference process.

\subsection{Video Action Detection}
\label{sec:action}
{\flushleft \bf Actor localization branch.} Benefit from preserving all keyframe tokens in Sec.~\ref{sec:token}, we can obtain multiple complete keyframe feature maps. We then up-sample or down-sample those keyframe feature maps for generating hierarchical features from the plain ViT. We then introduce a query-based actor localization head, inspired by Sparse R-CNN~\cite{sun2021sparse}, to detect actors in the keyframe.
The details of the localization head are provided in Appendix~\S~B.
Finally, the outputs of actor localization branch are $n$ prediction boxes in the keyframe and corresponding actor confidence scores.

{\flushleft \bf Action classification branch.}
Different from conventional feature extraction, EVAD produces $M$ discrete video tokens. We need to restore the spatiotemporal structure of video feature map and then can perform location-related operations such as RoIAlign. We initialize a blank feature map shaped as $(T/2, H/16, W/16)$, fill the preserved tokens into this feature map according to their corresponding spatiotemporal positions, and pad the rest with zeros.

Next, we use the boxes generated by the localization branch to extract actor RoI features via 3D RoIAlign for subsequent action prediction. Due to actor movement or camera translation, actor spatial position is changed across frames, and using the keyframe box for 3D RoIAlign cannot obtain partial actor feature deviated from the box. Directly extending the scope of the box to cover whole motion trajectory might harm actor feature representation by introducing background or other interfering information. Nevertheless, in EVAD feature extraction stage, the interfering ones in the video are progressively eliminated, thus we can properly extend the box scope to add the deviated feature. The results in Sec.~\ref{sec:ablation} demonstrate the ability of EVAD to cover large motion with extended boxes, while its counterpart ViTs impairs actor native feature representation.

We observe that directly applying token pruning methods of vision classification can greatly reduce the number of tokens involved in the calculations, but have a negative impact on final detection performance. Video action detection requires localizing and classifying the actions of all actors, but token pruning algorithms lead to incoherent actor features in space-time. As shown in Fig.~\ref{fig:vis_tp}, the static appearance of people sitting in the chairs is not retained completely in every frame. In the encoder, pair-wise self-attention is capable of modeling global dependency among tokens. The semantic information of the dropout tokens within the actor regions can be incorporated into some preserved tokens. Thus, we can recover the removed actor features from the preserved video tokens. To this end, we design a context refinement decoder to refine actor spatiotemporal representation. Concretely, we concatenate $n$ actor RoI features with $M$ video tokens and feed them into a sequentially-stacked transformer decoder consisting of MHSA and FFN. Guiding by the preserved tokens, actor features can enrich themselves with actor representation and motion information from other frames. Also, without token pruning, the decoder can be used as a kind of relational modeling modules~\cite{sun2018actor,DBLP:conf/cvpr/ZhangTHS19,girdhar2019video,DBLP:conf/eccv/TangXMPL20,pan2021actor} to capture inter-actor and actor-context interaction information. The actor feature refining and relational modeling capabilities of our decoder will be ablated in Sec.~\ref{sec:ablation}. 

The $n$ refined actor features output by the decoder are retrieved and passed through a classification layer to make final action prediction.

\section{Experiments}
\label{sec:exp}

\subsection{Experimental Setup}
{\flushleft \bf Datasets.} We evaluate our EVAD on three common datasets for video action detection: AVA~\cite{gu2018ava}, UCF101-24~\cite{soomro2012ucf101} and JHMDB~\cite{jhuang2013towards}. AVA is a large-scale benchmark and contains 299 15-minute videos, divided into 211k training clips and 57k validation clips. The videos are annotated at 1FPS for boxes and labels. Following the standard evaluation protocol, we report our performance on 60 common action classes. UCF101-24 is a subset of UCF101. It includes 3,207 videos from 24 sports classes. Each video contains a single action class. Following the common practice, we report the performance on split-1. JHMDB contains 928 trimmed videos from 21 action classes. We report the average results over all three splits.

{\flushleft \bf Evaluation criteria.} We evaluate the performances with frame level mean Average Precision (mAP) under IoU threshold of 0.5 without multiple scales and flips for fair comparisons. We measure the throughput on a single A100 GPU with a batch size of 8 to estimate the average number of images that can be processed in one second. We specify a sample of video clip containing 16 frame images by default. 

Our implementation details are described in Appendix~\S~B.

\subsection{Ablation Studies}
\label{sec:ablation}
We conduct in-depth ablation studies to investigate the effectiveness of our design in EVAD. All results are reported on AVA v2.2 with a 16-frame VideoMAE ViT-B backbone pre-trained and fine-tuned on Kinetics-400~\cite{kay2017kinetics}.

\setlength{\tabcolsep}{2pt}
\begin{table*}
\small
\centering
    \hspace{-3em}
    \begin{minipage}[t]{0.29\linewidth}
        \centering
        \begin{subtable}[t]{0.85\linewidth}
            \begin{tabular}{lx{10}x{25}x{28}}
            cls. head & $\rho$ & mAP & \#param\\
            \shline linear (baseline) & 1.0 & $26.4$ & $169$M \\
            WOO~\cite{chen2021watch} & 1.0 & $29.1$ & $314$M \\
            \rowcolor{baselinecolor} our decoder & 1.0 & $30.5$ & $185$M \\
            \hline linear & 0.7 & $22.5$ & $169$M \\
            \rowcolor{baselinecolor} our decoder & 0.7 & $30.5$ & $185$M \\
            \end{tabular}
            \caption{\textbf{Classification branch}}
            \label{tab:decoder}
        \end{subtable}
    \end{minipage}
    \hspace{1mm}
    \begin{minipage}[t]{0.22\linewidth}
        \centering
        \begin{subtable}[t]{0.85\linewidth}
            \begin{tabular}{lx{25}x{28}}
            case & mAP & GFLOPs\\
            \shline $\rho$=1.0 & $30.5$ & $223.8$ \\
            \hline random & $29.9$ & $158.9$ \\
            CLS token & $30.5$ & $134.2$ \\
            \rowcolor{baselinecolor} GAP & $30.5$ & $134.2$ \\
            & \\
            \end{tabular}
            % \vspace{1mm}
            \caption{\textbf{Pruning strategy}}
            \label{tab:type}
        \end{subtable}
    \end{minipage}
    \hspace{1mm}
    \begin{minipage}[t]{0.22\linewidth}
        \centering
        \begin{subtable}[t]{0.75\linewidth}
            \begin{tabular}{lx{25}}
            case & mAP\\
            \rowcolor{baselinecolor} \shline keyframe & $30.5$\\
            unified pruning & $30.0$\\
            & \\
            & \\
            & \\
            \end{tabular}
            \caption{\textbf{Keyframe}}
            \label{tab:key_tokens}
        \end{subtable}
    \end{minipage}
    \hspace{1mm}
    \begin{minipage}[t]{0.22\linewidth}
        \centering
        \begin{subtable}[t]{1.0\linewidth}
            \begin{tabular}{lx{25}x{28}}
            case & mAP & GFLOPs\\
            \rowcolor{baselinecolor} \shline preserved tokens & $30.5$ & $134.2$ \\
            all tokens & $30.2$ & $156.1$ \\
            \\
            \\
            \\
            \end{tabular}
            \caption{\textbf{Decoder input}}
            \label{tab:decoder_input}
            \vspace{1mm}
        \end{subtable}
    \end{minipage} \\
    \hspace{-2em}
    \begin{minipage}[t]{0.32\linewidth}
        \centering
        \begin{subtable}[t]{0.75\linewidth}
            \begin{tabular}{lx{18}x{18}x{18}x{18}x{18}}
            depth & 1 & 2 & 4 & 6 & 8\\
            \shline mAP & $27.2$ & $28.0$ & $29.7$ & \cellcolor{baselinecolor} $30.5$ & $30.3$\\
            \\
            \end{tabular}
            \caption{\textbf{Decoder depth}}
            \label{tab:depth}
        \end{subtable}
    \end{minipage}
    \hspace{1mm}
    \begin{minipage}[t]{0.34\linewidth}
        \centering
        \begin{subtable}[t]{1.\linewidth}
            \begin{tabular}{lx{23}x{32}x{32}x{32}}
            extend (x,y) & $(0,0)$ & $(0.2,0.1)$ & $(0.4,0.2)$ & $(0.6,0.3)$\\
            \shline $\rho$=1.0 & $30.5$ & $30.5$ & $30.2$ & $30.0$\\
            $\rho$=0.7 & $30.5$ & $30.5$ & \cellcolor{baselinecolor} $30.7$ & $30.4$ \\
            \end{tabular}
            \caption{\textbf{RoI extension}}
            \label{tab:extend}
        \end{subtable}
    \end{minipage}
    \hspace{1mm}
    \begin{minipage}[t]{0.33\linewidth}
        \centering
        \begin{subtable}[t]{0.75\linewidth}
            \begin{tabular}{lx{18}x{18}x{18}x{18}x{18}}
            ${w}_{kf}$ & $1$ & $2$ & $4$ & $6$ & $8$\\
            \shline mAP & $31.4$ & $31.7$ & \cellcolor{baselinecolor} $31.8$ & $31.7$ & $31.4$\\
            \\
            \end{tabular}
            \caption{\textbf{Attentiveness enhancement}}
            \label{tab:weight}
        \end{subtable}
    \end{minipage}
    \vspace{-2mm}
    \caption{\textbf{Ablation experiments on AVA v2.2.} All models here adopt 16-frame vanilla ViT-B as backbone and keep rate $\rho$ is set to 0.7 as the default setting unless otherwise specified.}
    \vspace{-1em}
\end{table*} 
\setlength{\tabcolsep}{1.4pt}

\begin{figure}[t!]
\centering
\includegraphics[width=0.95\linewidth]{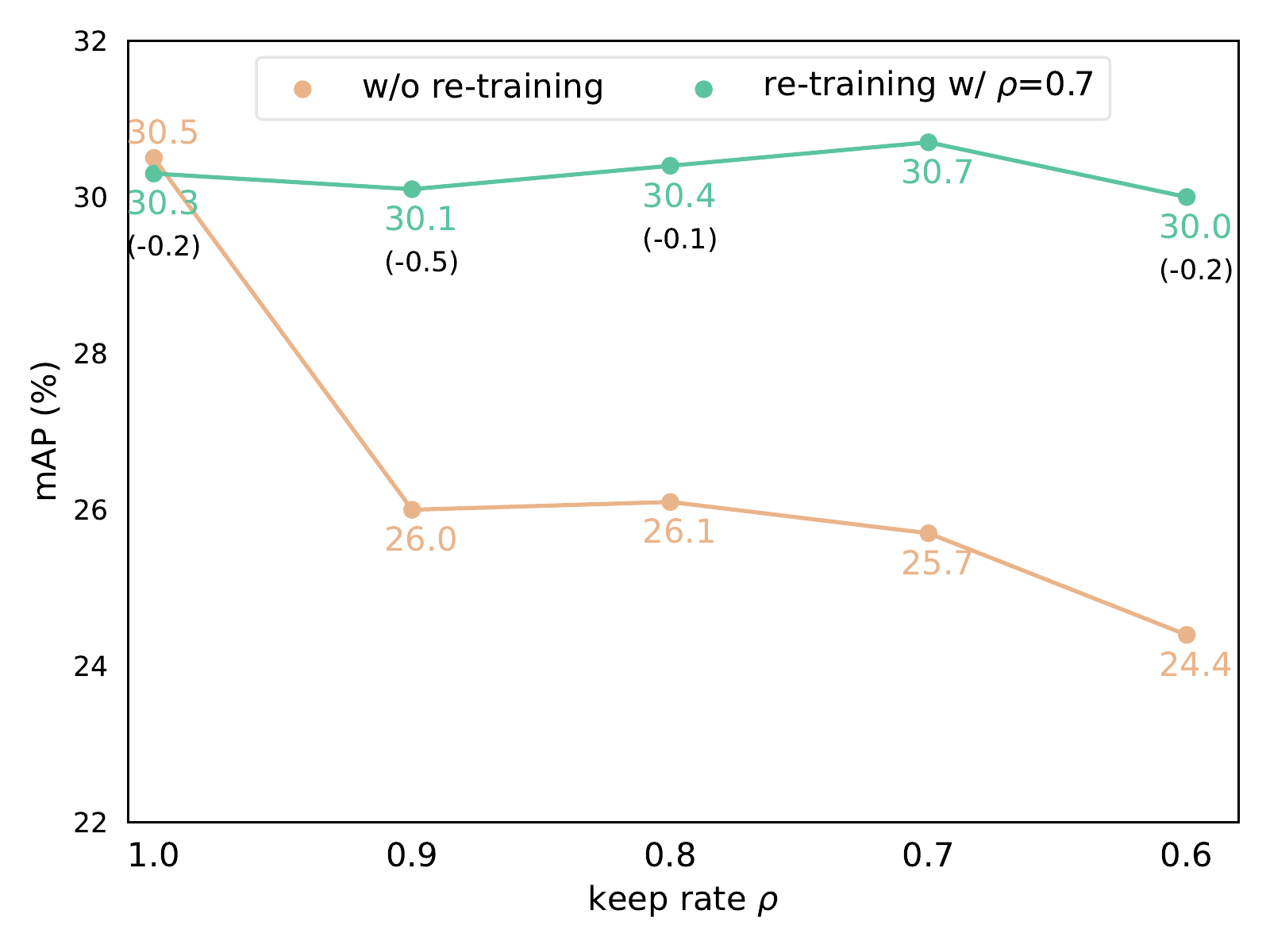}
\vspace{-3mm}
\caption{\textbf{Comparison on the performance of EVAD without re-training and re-training with keep rate $\rho$=0.7.} The numbers in `()' indicate the gap of the corresponding values to re-training with the target keep rate.}
\label{fig:infer}
\vspace{-2mm}
\end{figure}

{\flushleft \bf Inference w/o re-training.}
EVAD does not add any additional learnable parameters for token pruning. Thus, we directly add our keyframe-centric token pruning module on the off-the-shelf pre-trained ViTs and show the performance of the models with different token keep rates in Fig.~\ref{fig:infer}. We observe that directly applying token pruning negatively affects the model performance, and we speculate that the model needs re-training to adapt to a dynamic number of tokens at different layers.
Then, we incorporate our token pruning in the training process, and retrain EVAD with token keep rate of 0.7 and infer on multiple keep rates.
As shown in Fig.~\ref{fig:infer}, the result on each keep rate is nearly comparable to the re-training performance with the target keep rate. This illustrates that our token pruning module allows to adapt to different levels of computation costs after re-training once.

{\flushleft \bf Classification branch.}
To illustrate the necessity of refining actor spatiotemporal features when using token dropout for efficient action detection, we conduct two groups of experiments, i.e., the encoder with keyframe-centric token pruning ($\rho$=0.7) and without token pruning (keep rate $\rho$=1.0), and examine the effect of context refinement decoder for actor feature refinement and relational modeling. Without token pruning, the decoder is used to conduct actor-context modeling to capture action interaction information. It enables +4.1 mAP gains over the baseline and outperforms the well-designed WOO head with fewer parameters. When token pruning is applied, partial actor features are eliminated during the pruning process, and the linear model is much worse than the baseline. Nevertheless, the strong feature refining capability of decoder compensates this gap and achieves a comparable performance with $\rho$=1.0.

{\flushleft \bf Localization branch.} To explore the localization capability of EVAD and its impact on the overall performance, we directly use the boxes from an off-the-shelf detector~\cite{DBLP:conf/nips/RenHGS15}, which is commonly adopted in previous two-stage pipeline~\cite{DBLP:conf/iccv/Feichtenhofer0M19,DBLP:conf/cvpr/WuF0HKG19,DBLP:conf/eccv/TangXMPL20}. Our results shows that though utilizing the off-the-shelf detector can lead to an increase of +1.7 mAP, this comes at the cost of additional computational complexity, such as 246 GFLOPs for Faster R-CNN, and results in a more complex pipeline. The localization branch in our EVAD is lightweight with the cost of only 13.5 GFLOPs, and enjoys a simple end-to-end manner.

{\flushleft \bf Pruning strategy.} We then ablate different token pruning strategies with $\rho$=0.7. (1) Random masking performs a random sampling of non-keyframe tokens before entering the encoder. (2) Token pruning based on CLS token, as in~\cite{liang2022not}, uses a learnable CLS token pre-trained on video recognition~\cite{kay2017kinetics} responsible for calculating the importance of non-keyframe tokens. (3) GAP denotes pruning guided by global average pooling of all token attentive values.
As shown in Table~\ref{tab:type}, both attention-based approaches outperform random masking and achieve comparable performance with $\rho$=1.0, demonstrating that those models have the ability to preserve complete action semantics. We choose the latter with higher flexibility as the default setting.

{\flushleft \bf Keyframe.} In Sec~\ref{sec:token}, we emphasize the necessity of preserving all keyframe tokens. In Table~\ref{tab:key_tokens}, we compare the two strategies of keeping all keyframe tokens and unified pruning by treating keyframe tokens as normal tokens, the latter discarding some important tokens in the keyframe and leading to performance degradation.

{\flushleft \bf Decoder input.} To verify if some discarded tokens are important for action detection, we save those discarded ones when performing token pruning and feed all tokens into the decoder. As shown in Table~\ref{tab:decoder_input}, using all tokens does not improve the performance, indicating that 
the tokens preserved by the model contain sufficient action semantics.

{\flushleft \bf Decoder depth.} The results in Table~\ref{tab:depth} show that stacking 6 layers in the decoder has the highest mAP, and more than 6 layers cause performance degradation due to over-fitting, so we use depth=6 as the default setting.

\begin{figure}[t]
\centering
\includegraphics[width=1.\linewidth]{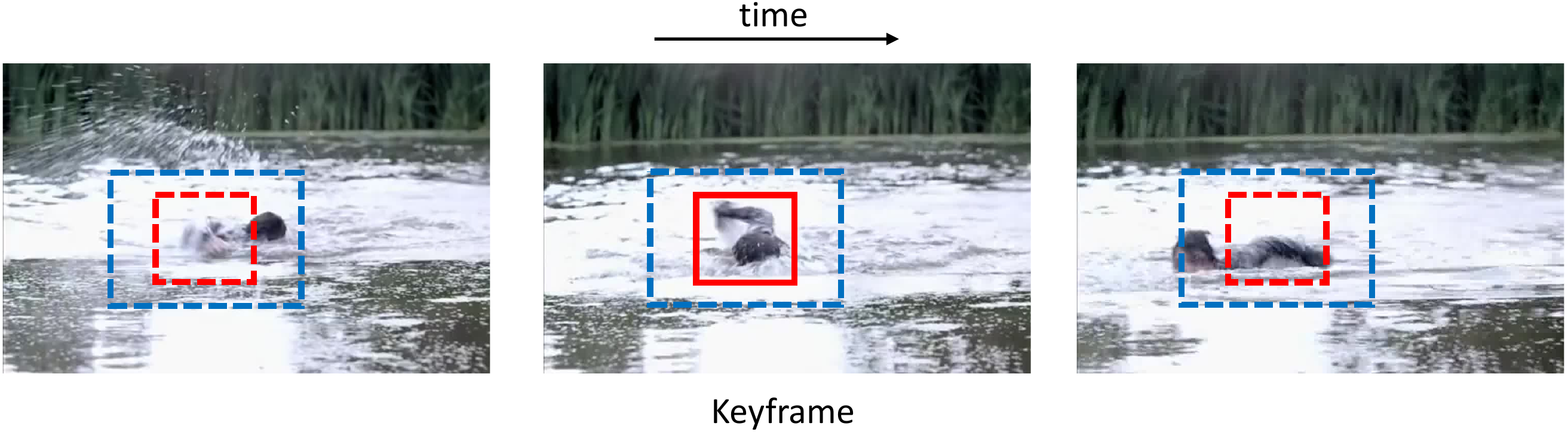}
\vspace{-2mm}
\caption{\textbf{RoI extension illustration.} The \textcolor[RGB]{234,51,35}{red} boxes denote the prediction box in the keyframe and are copied to adjacent frames. The \textcolor[RGB]{47,110,186}{blue} boxes denote the extended boxes to capture large motion.}
\label{fig:box_extend}
\vspace{-2mm}
\end{figure}

{\flushleft \bf RoI extension.} Due to human large motion, the box copied from the keyframe cannot cover the whole motion trajectory. Intuitively, we can solve it by properly extending the box scope, as shown in Fig.~\ref{fig:box_extend}, extending the scope of the red box in the keyframe to cover the swimming person.
In Table~\ref{tab:extend}, the model without token pruning shows a decreasing trend in performance as the proportion of box extending increases. Instead, the token pruning method achieves the best performance when the extended ratio reaches (0.4,0.2), where the box is $1.68\times$ of the original box, in which our pruning mechanism eliminates the interference information introduced by the extension.
It is worth mentioning that our method increases 15.7\% AP on `swim' category. We observe that large displacements frequently happen in `swim', and our token pruning combined with RoI extension can alleviate the effect of human movements to some extent.

{\flushleft \bf Attentiveness enhancement.} In Table~\ref{tab:keep_288}, there is a significant decrease in performance when the $\rho$ is reduced from 0.7 to 0.6. It indicates that the number of tokens retained at $\rho$=0.6 is insufficient to contain complete information. In Sec.~\ref{sec:token}, we analyze the portion of preserved tokens only has high responses to non-keyframes, which could be redundant to the current sample. 
Thus, we conduct experiments on $\rho$=0.6 to explore the feasibility of further reducing the number of tokens in Table~\ref{tab:weight}. We ablate the attention weight of keyframe queries, where ${w}_{kf}$=4 yields the best result. It shows that the keyframe plays a greater role in identifying the importance of non-keyframe tokens, leading to a lower redundancy of preserved tokens.

\begin{figure}[t!]
\centering
\includegraphics[width=0.8\linewidth]{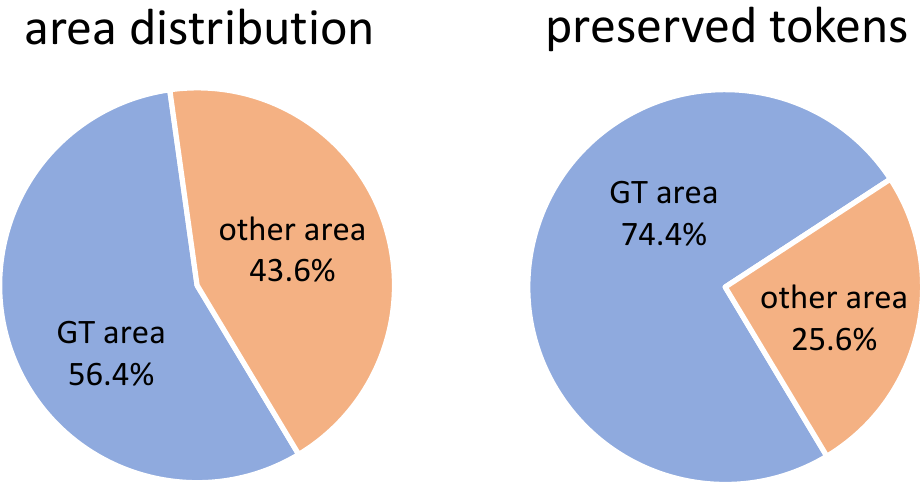}
\vspace{-1mm}
\caption{\textbf{Study on the relationship between the preserved tokens and GT bounding boxes.} `GT area' denotes the union of all ground-truth bounding boxes in a single sample, and `preserved tokens' in `GT area' represents the proportion of preserved tokens inside the GT area. The results are averaged over AVA validation set.}
\label{fig:gt}
\vspace{-3mm}
\end{figure}

{\flushleft \bf Discussion on the preserved tokens.} To explore what the preserved tokens correspond to, we compute the ratio of the preserved tokens inside the GT area on AVA validation set, as shown in Fig.~\ref{fig:gt}. We observe that the average GT area is 56.4\% of the input image, and 74.4\% of the preserved tokens belong to the GT area (56.4\%). This evidence indicates that, the proposed token pruning strategy allows our EVAD to focus on tokens in the GT area even without explicit box supervision.

\subsection{Efficiency Analysis}
\label{sec:analysis}
To verify the efficiency of EVAD, we measure several facets of computation, inference speed and training cost at multiple resolutions.

First, we compare saved computation and improved inference speed with different keep rates, measured by GFLOPs and throughput, respectively, as shown in Table~\ref{tab:keep}. When the keep rate $\rho$ is in the range of [0.7, 1.0), the performance is comparable to $\rho$=1.0, which indicates that high spatiotemporal redundancy exists in action detection and our method can effectively remove redundancy and maintain critical cues. The model with $\rho$=0.7 achieves the best performance-efficiency trade-off for both resolutions. 
For the short side of 224, the detection mAP improves 0.2\%, GFLOPs decrease by 40\%, and throughput increases by 38\%; for the short side of 288, mAP improves 0.2\%, GFLOPs decrease by 43\%, and throughput increases by 40\%. Further reducing the keep rate degrades the performance because the number of tokens kept is insufficient to contain complete semantic information.

Then, we fix the keep rate $\rho$ at 0.7 and compare the performance and efficiency gains of EVAD at higher resolutions, as shown in Table~\ref{table:higher}. We observe that: (1) EVAD with $\rho$=0.7 can obtain comparable performance to the model with $\rho$=1.0 at the same resolution with significantly lower computation and faster inference speed. (2) EVAD from a higher resolution provides a stable performance boost with relatively lower computation and faster speed. For instance, EVAD can improve performance by 1.1\% when increasing the resolution from 256 to 288, while GFLOPs decrease by 22\% and throughput increases by 10\%.

\setlength{\tabcolsep}{2pt}
\begin{table}[t]
\small
\centering
    \begin{center}
    \begin{subtable}[t]{0.85\linewidth}
        \centering
        \begin{tabular}{x{40}x{30}x{50}x{50}}
        \shline keep rate $\rho$ & mAP & GFLOPs & throughput\\ 
        \shline 1.0 & $30.5$ & $223.8$ & $298$ \\
        \hline $0.9$ & $30.6$ & $186.3$ \textcolor[RGB]{0,176,80}{(-$17\%$)} & $328$ \textcolor[RGB]{0,176,80}{(+$10\%$)} \\
        $0.8$ & $30.5$ & $157.0$ \textcolor[RGB]{0,176,80}{(-$30\%$)} & $377$ \textcolor[RGB]{0,176,80}{(+$27\%$)} \\
        \rowcolor{baselinecolor} $0.7$ & $30.7$ & $134.2$ \textcolor[RGB]{0,176,80}{(-$40\%$)} & $411$ \textcolor[RGB]{0,176,80}{(+$38\%$)} \\
        $0.6$ & $30.2$ & $116.3$ \textcolor[RGB]{0,176,80}{(-$48\%$)} & $427$ \textcolor[RGB]{0,176,80}{(+$43\%$)} \\
        \shline
        \end{tabular}
        \caption{Comparison on the resolution of short side 224.}
        \vspace{2mm}
        \label{tab:keep_224}
    \end{subtable}
    \end{center}
    \vspace{-2mm}
    \begin{subtable}[t]{0.85\linewidth}
        \centering
        \begin{tabular}{x{40}x{30}x{50}x{50}}
        \shline keep rate $\rho$ & mAP & GFLOPs & throughput\\ 
        \shline 1.0 & $32.1$ & $424.5$ & $171$ \\
        \hline $0.9$ & $32.2$ & $346.3$ \textcolor[RGB]{0,176,80}{(-$18\%$)} & $179$ \textcolor[RGB]{0,176,80}{(+$5\%$)} \\
        $0.8$ & $32.1$ & $287.4$ \textcolor[RGB]{0,176,80}{(-$32\%$)} & $216$ \textcolor[RGB]{0,176,80}{(+$26\%$)} \\
        \rowcolor{baselinecolor} $0.7$ & $32.3$ & $242.8$ \textcolor[RGB]{0,176,80}{(-$43\%$)} & $240$ \textcolor[RGB]{0,176,80}{(+$40\%$)} \\
        $0.6$ & $31.4$ & $208.9$ \textcolor[RGB]{0,176,80}{(-$51\%$)} & $263$ \textcolor[RGB]{0,176,80}{(+$54\%$)} \\
        \shline
        \end{tabular}
        \caption{Comparison on the resolution of short side 288.}
        \label{tab:keep_288}
    \end{subtable}
    \vspace{-2mm}
    \caption{\textbf{Comparison on the performance and efficiency variances of EVAD with different token keep rates.} The \textcolor[RGB]{0,176,80}{green} numbers indicate the gap of the corresponding values to EVAD with $\rho$=1.0.}
    \label{tab:keep}
\end{table} 
\setlength{\tabcolsep}{1.4pt}

\setlength{\tabcolsep}{2pt}
\begin{table}
\small
\begin{center}
\begin{tabular}{x{40}x{40}x{30}x{30}x{45}} 
\shline keep rate $\rho$ & image size & mAP & GFLOPs & throughput\\ 
\shline 1.0 & $224$ & $30.5$ & $223.8$ & $298$ \\
\rowcolor{baselinecolor} $0.7$ & $256$ & $31.4$ & $182.3$ & $334$ \\
1.0 & $256$ & $31.2$ & $311.9$ & $219$ \\
\rowcolor{baselinecolor} $0.7$ & $288$ & $32.3$ & $242.8$ & $240$ \\
1.0 & $288$ & $32.1$ & $424.5$ & $171$ \\
\shline
\end{tabular}
\vspace{-2mm}
\caption{\textbf{Comparison on the performance and efficiency improvements of EVAD with higher resolutions.}}
\label{table:higher}
\end{center}
\vspace{-3mm}
\end{table}
\setlength{\tabcolsep}{1.4pt}

\setlength{\tabcolsep}{2pt}
\begin{table}
\small
\begin{center}
\begin{tabular}{lx{35}x{30}x{35}x{40}}
\shline model & backbone & mAP & GFLOPs & throughput\\
\shline
 WOO~\cite{chen2021watch} & SFR101 & 28.3 & 252 & 147\\
 WOO~\cite{chen2021watch}$^\ddag$ & ViT-B & 30.0 & 378 & 176\\
 TubeR~\cite{zhao2022tuber} & CSN-152 & 33.6 & 240 & 64$^{*}$\\
 VideoMAE~\cite{tong2022videomae} & ViT-B & 31.8 & 180+246 & N/A\\
 VideoMAE~\cite{tong2022videomae} & ViT-L & 39.3 & 597+246 & N/A\\
 \hline $\mathbf{EVAD}$, $\rho$=0.7 & ViT-B & 32.3 & 243 & 240\\
 $\mathbf{EVAD}$, $\rho$=0.7 & ViT-L & 39.7 & 737 & 153\\
\shline
\vspace{-4mm}
\end{tabular}
\caption{\textbf{Comparison on the performance and efficiency of EVAD and other state-of-the-art methods.} $^\ddag$ denotes the results re-implemented by us for fair comparison. *: the code provided by TubeR does not include the implementation of its long-term context head, so the actual throughput will be less than we measured. `N/A' indicates that the two-stage pipeline utilized in VideoMAE is not applicable to measure throughput.}
\label{table:compare}
\end{center}
\vspace{-2em}
\end{table}
\setlength{\tabcolsep}{1.4pt}

\begin{figure}[t]
\centering
\includegraphics[width=0.95\linewidth]{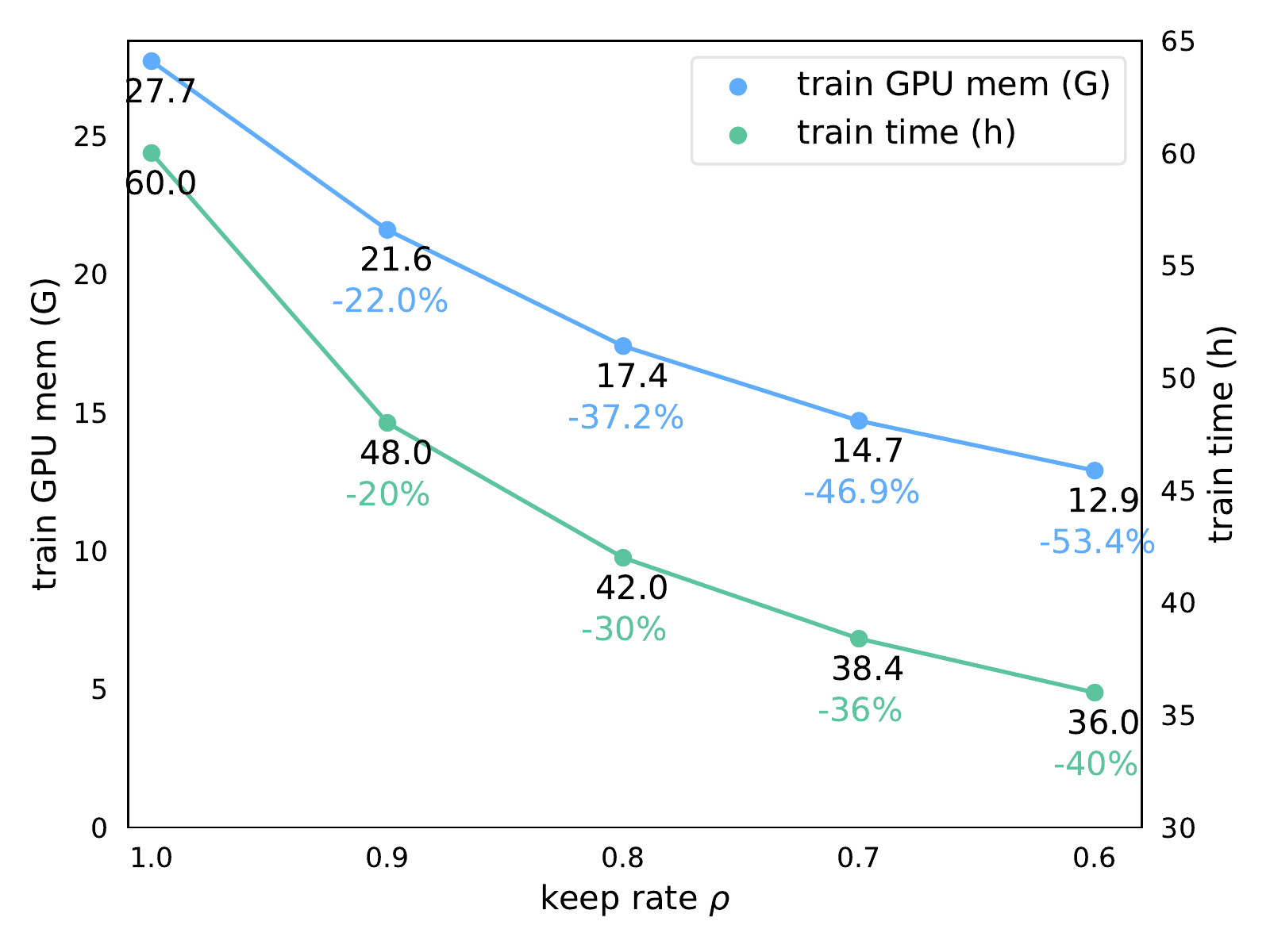}
\vspace{-3mm}
\caption{\textbf{Comparison on GPU memory and training time required of EVAD training with different token keep rates.} The \textcolor[RGB]{96,172,252}{blue} and \textcolor[RGB]{91,196,159}{green} numbers indicate the gap of the corresponding values to EVAD with $\rho$=1.0.}
\label{fig:eff_train}
\vspace{-4mm}
\end{figure}

\setlength{\tabcolsep}{2pt}
\begin{table*}
\small
\begin{center}
    \begin{minipage}[t]{0.57\linewidth}
        \centering
        \begin{subtable}[t]{1.\linewidth}
            \begin{tabular}{lcccccc} 
                \shline model & e2e & $T \times \tau$ & backbone & pre-train & GFLOPs & mAP \\
                \hline SlowFast~\cite{DBLP:conf/iccv/Feichtenhofer0M19} & \XSolidBrush & $32\times 2$ & SF-R101-NL & K600 & 393 & $29.0$ \\
                ACAR-Net~\cite{pan2021actor} & \XSolidBrush & $32\times 2$ & SF-R101-NL & K600 & N/A & $31.4$ \\
                AIA~\cite{DBLP:conf/eccv/TangXMPL20} & \XSolidBrush & $32\times 2$ & SF-R101 & K700 & N/A & $32.3$ \\
                MeMViT~\cite{wu2022memvit} & \XSolidBrush & $16\times 4$ & MViTv2-S & K400 & 305 & $29.3$ \\
                MeMViT~\cite{wu2022memvit} & \XSolidBrush & $32\times 3$ & MViTv2-B & K700 & 458 & $34.4$ \\
                WOO~\cite{chen2021watch} & \Checkmark & $32\times 2$ & SF-R101-NL & K600 & 252 & $28.3$ \\
                WOO~\cite{chen2021watch}$^\ddag$ & \Checkmark & $16\times 4$ & ViT-B & K400 & 378 & $30.0$ \\
                TubeR~\cite{zhao2022tuber} & \Checkmark & $32\times 2$ & CSN-152 & IG+K400 & 240 & $33.6$ \\                MaskFeat{\scriptsize\textuparrow312}~\cite{wei2021masked} & \XSolidBrush & $40\times 3$ & MViTv2-L & K400 & 3074 & $37.5$ \\
                VideoMAE~\cite{tong2022videomae} & \XSolidBrush & $16\times 4$ & ViT-B & K400 & 426 & $31.8$ \\
                VideoMAE~\cite{tong2022videomae} & \XSolidBrush & $16\times 4$ & ViT-L & K700 & 843 & $39.3$ \\
                STMixer~\cite{wu2023stmixer} & \Checkmark & $16\times 4$ & ViT-L & K700 & N/A & $39.5$ \\
                \hline $\mathbf{EVAD}$, $\rho$=0.7 & \Checkmark & $16\times 4$ & ViT-B & K400 & $\textbf{243}$ & \textbf{32.3} \\
                $\mathbf{EVAD}$, $\rho$=0.7 & \Checkmark & $16\times 4$ & ViT-B$^\dag$ & K710+K400 & $\textbf{243}$ & \textbf{37.7} \\
                $\mathbf{EVAD}$, $\rho$=0.7 & \Checkmark & $16\times 4$ & ViT-L & K700 & $\textbf{737}$ & $\textbf{39.7}$ \\
            \shline
            \end{tabular}
            \caption{\textbf{Comparison on AVA v2.2.}}
            \label{table:ava_sota}
        \end{subtable}
    \end{minipage}
    \begin{minipage}[t]{0.42\linewidth}
        \centering
        \begin{subtable}[t]{1.\linewidth}
            \begin{tabular}{lccccc} 
                \shline model & e2e & $T \times \tau$ & backbone & JHMDB & UCF24 \\
                \hline ACT~\cite{kalogeiton2017action} & \Checkmark & $6\times 1$ & VGG & $65.7$ & $69.5$ \\
                MOC~\cite{DBLP:conf/eccv/LiW0W20}$^\ast$ & \Checkmark & $7\times 1$ & DLA32 & $70.8$ & $78.0$ \\
                AVA~\cite{gu2018ava}$^\ast$ & \XSolidBrush &$20\times 1$ & I3D-VGG & $73.3$ & $76.3$ \\
                ACRN~\cite{sun2018actor} & \Checkmark & $20\times 1$ & S3D-G & $77.9$ & - \\
                CA-RCNN~\cite{wu2020context} & \XSolidBrush & $32\times 2$ & R50-NL & $79.2$ & - \\
                YOWO~\cite{kopuklu2019you} & \Checkmark & $16\times 1$ & 3D-X101 & $80.4$ & $75.7$ \\
                WOO~\cite{chen2021watch} & \Checkmark & $32\times 2$ & SF-R101-NL & $80.5$ & - \\
                TubeR~\cite{zhao2022tuber}$^\ast$ & \Checkmark & $32\times 2$ & I3D & $80.7$ & $81.3$ \\
                TubeR~\cite{zhao2022tuber} & \Checkmark & $32\times 2$ & CSN-152 & - & $83.2$ \\
                AIA~\cite{DBLP:conf/eccv/TangXMPL20} & \XSolidBrush & $32\times 1$ & R50-C2D & - & $78.8$ \\
                ACAR-Net~\cite{pan2021actor} & \XSolidBrush & $32\times 1$ & SF-R50 & - & $84.3$ \\
                STMixer~\cite{wu2023stmixer} & \Checkmark & $32\times 2$ & SF-R101-NL & $86.7$ & $83.7$ \\
                \hline 
                baseline (Ours)  & \Checkmark & $16\times 4$ & ViT-B & $88.1$ & $83.9$ \\
                $\mathbf{EVAD}$, $\rho$=1.0 & \Checkmark & $16\times 4$ & ViT-B & $88.5$ & $84.9$ \\
                $\mathbf{EVAD}$, $\rho$=0.6 & \Checkmark & $16\times 4$ & ViT-B & $\mathbf{90.2}$ & $\mathbf{85.1}$ \\
                \shline
            \end{tabular}
            \caption{\textbf{Comparison on JHMDB and UCF101-24.}}
            \label{table:ucf_sota}
        \end{subtable}
    \end{minipage}
    \vspace{-1mm}
    \caption{\textbf{Comparison with the state-of-the-art on three benchmarks.} \Checkmark denotes an end-to-end approach using a unified backbone, and \XSolidBrush denotes the use of two separated backbones, one of which is Faster R-CNN-R101-FPN (246 GFLOPs~\cite{DBLP:conf/nips/RenHGS15}) to pre-compute person proposals. $^\ddag$ denotes the results re-implemented by us for fair comparison. $T \times \tau$ refers to the frame number and corresponding sample rate. Methods marked with $^\ast$ leverage optical flow input. $^\dag$: ViT-B backbone pre-trained on K710 and fine-tuned on K400 from VideoMAE V2~\cite{wang2023videomae}.}
\end{center}
\vspace{-5mm}
\end{table*}

Next, we compare the performance and efficiency of EVAD with other state-of-the-art methods in Table~\ref{table:compare}. We first observe that vanilla ViT allows WOO to achieve faster inference speed compared to traditional CNN-based backbone, and our EVAD can further accelerate the inference process (240 img/s vs. 176 img/s). Moreover, with a larger ViT-L backbone, EVAD can maintain a comparable inference speed to WOO with a CNN backbone (153 img/s vs. 147 img/s), and obtain significant performance improvements (39.7 mAP vs. 28.3 mAP). Although the previous art TubeR has fewer GFLOPs, it works slowly due to its heavy CSN-152 backbone. Compared to VideoMAE, EVAD can achieve comparable performance with fewer GFLOPs and allow for real-time inference in an end-to-end manner.

Finally, we show the significant impacts of EVAD during the training phase. In Fig.~\ref{fig:eff_train}, as the keep rate decreases, GPU memory and training time consequently reduces during the EVAD training. When the keep rate $\rho$ is 0.7, we can save 47\% of GPU memory and 36\% of training time, while maintaining the similar performance as EVAD with $\rho$=1.0. 

In summary, EVAD can achieve significant improvements in saving computations, improving inference speed and reducing training costs. It can be treated as a simple yet efficient baseline applicable to many other action detection methods.

\subsection{Comparison with the state-of-the-art methods}
\label{sec:sota}
We compare our EVAD with the state-of-the-art methods on AVA v2.2 in Table~\ref{table:ava_sota}. The result on AVA v2.1 is provided in Appendix~\S~C. To the best of our knowledge, the prevailing best model on AVA is VideoMAE, a two-stage model that requires an off-the-shelf person detector to pre-compute person proposals. Using the same pre-trained backbone, we can obtain comparable performance to VideoMAE, with mAP of $32.3$ vs. $31.8$ on ViT-B and $39.7$ vs. $39.3$ on ViT-L, and surpass other two-stage models.
Compared to end-to-end models such as WOO and TubeR, we also have a significant performance gain, while benefiting from the structural nature of transformers and token pruning with a low keep rate, we have a faster inference speed than CNN-based models and are more friendly to real-time action detection. Moreover, we can achieve slightly better performance than STMixer. STMixer is the most recent end-to-end model that designs a decoder to sample discriminative features. Orthogonal to it, our EVAD is devised for efficient video feature extraction, and further combining the two may yield better detection performance.

To demonstrate the generalizability of EVAD, we further verify our model on JHMDB and UCF101-24. As shown in Table~\ref{table:ucf_sota}, we achieve state-of-the-art performance on both datasets with comparable improvement. Different from the experimental results on AVA, using the keep rate of 0.6 can lead to similar or better results than the model without token pruning on these two datasets. We consider the reason is that the scenarios of JHDMB and UCF01-24 are simpler and do not require complex relational modeling of actor-actor and actor-object, and hence can preserve a fewer number of tokens. In addition, the JHDMB dataset is small and using a lower keep rate for training might alleviate over-fitting and learn a better feature representation. We also provide the video-mAP results on both datasets in Appendix~\S~C.

\section{Conclusion and Future Work}
Motivated by expensive computational costs of transformers with a video sequence and high spatiotemporal redundancy in video action detection, we design an Efficient Video Action Detector (EVAD) by dropping out spatiotemporal tokens and refining scene context to enable efficient transformer-based action detection.
EVAD can achieve comparable performance to the vanilla ViT while considerably reducing computational costs and expediting inference speed, and it can reach the state-of-the-art on three popular action detection datasets. We hope that EVAD can serve as an efficient end-to-end baseline for future studies.

One limitation of our approach is that EVAD requires re-training once to take the benefits of reduced computations and faster inference from removing redundancy. A potential research approach is to explore transformer-adaptive token pruning algorithms. Moreover, we follow the end-to-end framework of WOO to verify the efficiency and effectiveness of EVAD, but WOO is still a `two-stage' pipeline, which sequentially conducts actor localization and action classification modules. In future work, we aim to integrate those two modules into a unified head, which can reduce the inference time of passing through the detector head and hence amplify the efficiency benefits of EVAD.

{\flushleft \bf Acknowledgements.} This work is supported by National Key R$\&$D Program of China (No. 2022ZD0160900), National Natural Science Foundation of China (No. 62076119, No. 61921006, No. 62072232)), Fundamental Research Funds for the Central Universities (No. 020214380091, No. 020214380099), and Collaborative Innovation Center of Novel Software Technology and Industrialization.

\clearpage
{\small
\bibliographystyle{ieee_fullname}
\bibliography{ref}

\begin{thebibliography}{10}\itemsep=-1pt

\bibitem{arnab2021vivit}
Anurag Arnab, Mostafa Dehghani, Georg Heigold, Chen Sun, Mario Lu{\v{c}}i{\'c},
  and Cordelia Schmid.
\newblock Vivit: A video vision transformer.
\newblock In {\em IEEE/CVF International Conference on Computer Vision}, 2021.

\bibitem{arnab2022beyond}
Anurag Arnab, Xuehan Xiong, Alexey Gritsenko, Rob Romijnders, Josip Djolonga,
  Mostafa Dehghani, Chen Sun, Mario Lu{\v{c}}i{\'c}, and Cordelia Schmid.
\newblock Beyond transfer learning: Co-finetuning for action localisation.
\newblock {\em arXiv preprint arXiv:2207.03807}, 2022.

\bibitem{bertasius2021space}
Gedas Bertasius, Heng Wang, and Lorenzo Torresani.
\newblock Is space-time attention all you need for video understanding?
\newblock In {\em International Conference on Machine Learning}, 2021.

\bibitem{bulat2021space}
Adrian Bulat, Juan~Manuel Perez~Rua, Swathikiran Sudhakaran, Brais Martinez,
  and Georgios Tzimiropoulos.
\newblock Space-time mixing attention for video transformer.
\newblock {\em Advances in Neural Information Processing Systems}, 2021.

\bibitem{carion2020end}
Nicolas Carion, Francisco Massa, Gabriel Synnaeve, Nicolas Usunier, Alexander
  Kirillov, and Sergey Zagoruyko.
\newblock End-to-end object detection with transformers.
\newblock In {\em European Conference on Computer Vision}, 2020.

\bibitem{chen2023cycleacr}
Lei Chen, Zhan Tong, Yibing Song, Gangshan Wu, and Limin Wang.
\newblock Cycleacr: Cycle modeling of actor-context relations for video action
  detection.
\newblock {\em arXiv preprint arXiv:2303.16118}, 2023.

\bibitem{chen2021watch}
Shoufa Chen, Peize Sun, Enze Xie, Chongjian Ge, Jiannan Wu, Lan Ma, Jiajun
  Shen, and Ping Luo.
\newblock Watch only once: An end-to-end video action detection framework.
\newblock In {\em IEEE/CVF International Conference on Computer Vision}, 2021.

\bibitem{dai2021up}
Zhigang Dai, Bolun Cai, Yugeng Lin, and Junying Chen.
\newblock Up-detr: Unsupervised pre-training for object detection with
  transformers.
\newblock In {\em IEEE/CVF Conference on Computer Vision and Pattern
  Recognition}, 2021.

\bibitem{dosovitskiy2020image}
Alexey Dosovitskiy, Lucas Beyer, Alexander Kolesnikov, Dirk Weissenborn,
  Xiaohua Zhai, Thomas Unterthiner, Mostafa Dehghani, Matthias Minderer, Georg
  Heigold, Sylvain Gelly, et~al.
\newblock An image is worth 16x16 words: Transformers for image recognition at
  scale.
\newblock {\em arXiv preprint arXiv:2010.11929}, 2020.

\bibitem{fan2021multiscale}
Haoqi Fan, Bo Xiong, Karttikeya Mangalam, Yanghao Li, Zhicheng Yan, Jitendra
  Malik, and Christoph Feichtenhofer.
\newblock Multiscale vision transformers.
\newblock In {\em IEEE/CVF International Conference on Computer Vision}, 2021.

\bibitem{fayyaz20213d}
Mohsen Fayyaz, Emad Bahrami, Ali Diba, Mehdi Noroozi, Ehsan Adeli, Luc
  Van~Gool, and Jurgen Gall.
\newblock 3d cnns with adaptive temporal feature resolutions.
\newblock In {\em IEEE/CVF Conference on Computer Vision and Pattern
  Recognition}, 2021.

\bibitem{fayyaz2022adaptive}
Mohsen Fayyaz, Soroush~Abbasi Koohpayegani, Farnoush~Rezaei Jafari, Sunando
  Sengupta, Hamid Reza~Vaezi Joze, Eric Sommerlade, Hamed Pirsiavash, and
  J{\"u}rgen Gall.
\newblock Adaptive token sampling for efficient vision transformers.
\newblock In {\em European Conference on Computer Vision}, 2022.

\bibitem{feichtenhofer2020x3d}
Christoph Feichtenhofer.
\newblock X3d: Expanding architectures for efficient video recognition.
\newblock In {\em IEEE/CVF Conference on Computer Vision and Pattern
  Recognition}, 2020.

\bibitem{DBLP:conf/iccv/Feichtenhofer0M19}
Christoph Feichtenhofer, Haoqi Fan, Jitendra Malik, and Kaiming He.
\newblock Slowfast networks for video recognition.
\newblock In {\em IEEE/CVF International Conference on Computer Vision}, 2019.

\bibitem{girdhar2019video}
Rohit Girdhar, Joao Carreira, Carl Doersch, and Andrew Zisserman.
\newblock Video action transformer network.
\newblock In {\em IEEE/CVF Conference on Computer Vision and Pattern
  Recognition}, 2019.

\bibitem{glorot2010understanding}
Xavier Glorot and Yoshua Bengio.
\newblock Understanding the difficulty of training deep feedforward neural
  networks.
\newblock In {\em Proceedings of the thirteenth international conference on
  artificial intelligence and statistics}, 2010.

\bibitem{gu2018ava}
Chunhui Gu, Chen Sun, David~A Ross, Carl Vondrick, Caroline Pantofaru, Yeqing
  Li, Sudheendra Vijayanarasimhan, George Toderici, Susanna Ricco, Rahul
  Sukthankar, et~al.
\newblock Ava: A video dataset of spatio-temporally localized atomic visual
  actions.
\newblock In {\em IEEE/CVF Conference on Computer Vision and Pattern
  Recognition}, 2018.

\bibitem{DBLP:conf/iccv/HeGDG17}
Kaiming He, Georgia Gkioxari, Piotr Doll{\'{a}}r, and Ross~B. Girshick.
\newblock Mask {R-CNN}.
\newblock In {\em IEEE/CVF International Conference on Computer Vision}, 2017.

\bibitem{heo2021rethinking}
Byeongho Heo, Sangdoo Yun, Dongyoon Han, Sanghyuk Chun, Junsuk Choe, and
  Seong~Joon Oh.
\newblock Rethinking spatial dimensions of vision transformers.
\newblock In {\em IEEE/CVF International Conference on Computer Vision}, 2021.

\bibitem{jhuang2013towards}
Hueihan Jhuang, Juergen Gall, Silvia Zuffi, Cordelia Schmid, and Michael~J
  Black.
\newblock Towards understanding action recognition.
\newblock In {\em IEEE/CVF International Conference on Computer Vision}, 2013.

\bibitem{jiang2021all}
Zi-Hang Jiang, Qibin Hou, Li Yuan, Daquan Zhou, Yujun Shi, Xiaojie Jin, Anran
  Wang, and Jiashi Feng.
\newblock All tokens matter: Token labeling for training better vision
  transformers.
\newblock {\em Advances in Neural Information Processing Systems}, 2021.

\bibitem{kalogeiton2017action}
Vicky Kalogeiton, Philippe Weinzaepfel, Vittorio Ferrari, and Cordelia Schmid.
\newblock Action tubelet detector for spatio-temporal action localization.
\newblock In {\em Proceedings of the IEEE International Conference on Computer
  Vision}, pages 4405--4413, 2017.

\bibitem{kay2017kinetics}
Will Kay, Joao Carreira, Karen Simonyan, Brian Zhang, Chloe Hillier, Sudheendra
  Vijayanarasimhan, Fabio Viola, Tim Green, Trevor Back, Paul Natsev, et~al.
\newblock The kinetics human action video dataset.
\newblock {\em arXiv preprint arXiv:1705.06950}, 2017.

\bibitem{kopuklu2019you}
Okan K{\"o}p{\"u}kl{\"u}, Xiangyu Wei, and Gerhard Rigoll.
\newblock You only watch once: A unified cnn architecture for real-time
  spatiotemporal action localization.
\newblock {\em arXiv preprint arXiv:1911.06644}, 2019.

\bibitem{korbar2019scsampler}
Bruno Korbar, Du Tran, and Lorenzo Torresani.
\newblock Scsampler: Sampling salient clips from video for efficient action
  recognition.
\newblock In {\em IEEE/CVF International Conference on Computer Vision}, 2019.

\bibitem{kumar2022end}
Akash Kumar and Yogesh~Singh Rawat.
\newblock End-to-end semi-supervised learning for video action detection.
\newblock In {\em IEEE/CVF Conference on Computer Vision and Pattern
  Recognition}, 2022.

\bibitem{li2021multisports}
Yixuan Li, Lei Chen, Runyu He, Zhenzhi Wang, Gangshan Wu, and Limin Wang.
\newblock Multisports: A multi-person video dataset of spatio-temporally
  localized sports actions.
\newblock In {\em IEEE/CVF International Conference on Computer Vision}, 2021.

\bibitem{DBLP:conf/eccv/LiW0W20}
Yixuan Li, Zixu Wang, Limin Wang, and Gangshan Wu.
\newblock Actions as moving points.
\newblock In {\em European Conference on Computer Vision}, 2020.

\bibitem{liang2022not}
Youwei Liang, Chongjian Ge, Zhan Tong, Yibing Song, Jue Wang, and Pengtao Xie.
\newblock Not all patches are what you need: Expediting vision transformers via
  token reorganizations.
\newblock In {\em International Conference on Learning Representations}, 2022.

\bibitem{ocsampler}
Jintao Lin, Haodong Duan, Kai Chen, Dahua Lin, and Limin Wang.
\newblock Ocsampler: Compressing videos to one clip with single-step sampling.
\newblock In {\em IEEE/CVF Conference on Computer Vision and Pattern
  Recognition}, pages 13884--13893, 2022.

\bibitem{lin2017feature}
Tsung-Yi Lin, Piotr Doll{\'a}r, Ross Girshick, Kaiming He, Bharath Hariharan,
  and Serge Belongie.
\newblock Feature pyramid networks for object detection.
\newblock In {\em IEEE/CVF Conference on Computer Vision and Pattern
  Recognition}, 2017.

\bibitem{liu2022video}
Ze Liu, Jia Ning, Yue Cao, Yixuan Wei, Zheng Zhang, Stephen Lin, and Han Hu.
\newblock Video swin transformer.
\newblock In {\em IEEE/CVF Conference on Computer Vision and Pattern
  Recognition}, 2022.

\bibitem{loshchilov2017decoupled}
Ilya Loshchilov and Frank Hutter.
\newblock Decoupled weight decay regularization.
\newblock {\em arXiv preprint arXiv:1711.05101}, 2017.

\bibitem{marin2021token}
Dmitrii Marin, Jen-Hao~Rick Chang, Anurag Ranjan, Anish Prabhu, Mohammad
  Rastegari, and Oncel Tuzel.
\newblock Token pooling in vision transformers.
\newblock {\em arXiv preprint arXiv:2110.03860}, 2021.

\bibitem{pan2021actor}
Junting Pan, Siyu Chen, Mike~Zheng Shou, Yu Liu, Jing Shao, and Hongsheng Li.
\newblock Actor-context-actor relation network for spatio-temporal action
  localization.
\newblock In {\em IEEE/CVF Conference on Computer Vision and Pattern
  Recognition}, 2021.

\bibitem{park2022k}
Seong~Hyeon Park, Jihoon Tack, Byeongho Heo, Jung-Woo Ha, and Jinwoo Shin.
\newblock K-centered patch sampling for efficient video recognition.
\newblock In {\em European Conference on Computer Vision}, 2022.

\bibitem{qing2022mar}
Zhiwu Qing, Shiwei Zhang, Ziyuan Huang, Xiang Wang, Yuehuan Wang, Yiliang Lv,
  Changxin Gao, and Nong Sang.
\newblock Mar: Masked autoencoders for efficient action recognition.
\newblock {\em arXiv preprint arXiv:2207.11660}, 2022.

\bibitem{rao2021dynamicvit}
Yongming Rao, Wenliang Zhao, Benlin Liu, Jiwen Lu, Jie Zhou, and Cho-Jui Hsieh.
\newblock Dynamicvit: Efficient vision transformers with dynamic token
  sparsification.
\newblock {\em Advances in Neural Information Processing Systems}, 2021.

\bibitem{DBLP:conf/nips/RenHGS15}
Shaoqing Ren, Kaiming He, Ross~B. Girshick, and Jian Sun.
\newblock Faster {R-CNN:} towards real-time object detection with region
  proposal networks.
\newblock In {\em Advances in Neural Information Processing Systems}, 2015.

\bibitem{renggli2022learning}
Cedric Renggli, Andr{\'e}~Susano Pinto, Neil Houlsby, Basil Mustafa, Joan
  Puigcerver, and Carlos Riquelme.
\newblock Learning to merge tokens in vision transformers.
\newblock {\em arXiv preprint arXiv:2202.12015}, 2022.

\bibitem{ryoo2021tokenlearner}
Michael~S. Ryoo, AJ Piergiovanni, Anurag Arnab, Mostafa Dehghani, and Anelia
  Angelova.
\newblock Tokenlearner: Adaptive space-time tokenization for videos.
\newblock In {\em Advances in Neural Information Processing Systems}, 2021.

\bibitem{song2021vidt}
Hwanjun Song, Deqing Sun, Sanghyuk Chun, Varun Jampani, Dongyoon Han, Byeongho
  Heo, Wonjae Kim, and Ming-Hsuan Yang.
\newblock Vidt: An efficient and effective fully transformer-based object
  detector.
\newblock {\em arXiv preprint arXiv:2110.03921}, 2021.

\bibitem{soomro2012ucf101}
Khurram Soomro, Amir~Roshan Zamir, and Mubarak Shah.
\newblock Ucf101: A dataset of 101 human actions classes from videos in the
  wild.
\newblock {\em arXiv preprint arXiv:1212.0402}, 2012.

\bibitem{sun2018actor}
Chen Sun, Abhinav Shrivastava, Carl Vondrick, Kevin Murphy, Rahul Sukthankar,
  and Cordelia Schmid.
\newblock Actor-centric relation network.
\newblock In {\em European Conference on Computer Vision}, 2018.

\bibitem{sun2021sparse}
Peize Sun, Rufeng Zhang, Yi Jiang, Tao Kong, Chenfeng Xu, Wei Zhan, Masayoshi
  Tomizuka, Lei Li, Zehuan Yuan, Changhu Wang, et~al.
\newblock Sparse r-cnn: End-to-end object detection with learnable proposals.
\newblock In {\em IEEE/CVF Conference on Computer Vision and Pattern
  Recognition}, 2021.

\bibitem{DBLP:conf/eccv/TangXMPL20}
Jiajun Tang, Jin Xia, Xinzhi Mu, Bo Pang, and Cewu Lu.
\newblock Asynchronous interaction aggregation for action detection.
\newblock In {\em European Conference on Computer Vision}, 2020.

\bibitem{tong2022videomae}
Zhan Tong, Yibing Song, Jue Wang, and Limin Wang.
\newblock Video{MAE}: Masked autoencoders are data-efficient learners for
  self-supervised video pre-training.
\newblock In {\em Advances in Neural Information Processing Systems}, 2022.

\bibitem{touvron2021training}
Hugo Touvron, Matthieu Cord, Matthijs Douze, Francisco Massa, Alexandre
  Sablayrolles, and Herv{\'e} J{\'e}gou.
\newblock Training data-efficient image transformers \& distillation through
  attention.
\newblock In {\em International Conference on Machine Learning}, 2021.

\bibitem{wang2022efficient}
Junke Wang, Xitong Yang, Hengduo Li, Li Liu, Zuxuan Wu, and Yu-Gang Jiang.
\newblock Efficient video transformers with spatial-temporal token selection.
\newblock In {\em European Conference on Computer Vision}, 2022.

\bibitem{wang2023videomae}
Limin Wang, Bingkun Huang, Zhiyu Zhao, Zhan Tong, Yinan He, Yi Wang, Yali Wang,
  and Yu Qiao.
\newblock Videomae v2: Scaling video masked autoencoders with dual masking.
\newblock In {\em IEEE/CVF Conference on Computer Vision and Pattern
  Recognition}, 2023.

\bibitem{wang2021adaptive}
Yulin Wang, Zhaoxi Chen, Haojun Jiang, Shiji Song, Yizeng Han, and Gao Huang.
\newblock Adaptive focus for efficient video recognition.
\newblock In {\em IEEE/CVF International Conference on Computer Vision}, 2021.

\bibitem{wei2021masked}
Chen Wei, Haoqi Fan, Saining Xie, Chao-Yuan Wu, Alan Yuille, and Christoph
  Feichtenhofer.
\newblock Masked feature prediction for self-supervised visual pre-training.
\newblock In {\em IEEE/CVF Conference on Computer Vision and Pattern
  Recognition}, 2022.

\bibitem{DBLP:conf/cvpr/WuF0HKG19}
Chao{-}Yuan Wu, Christoph Feichtenhofer, Haoqi Fan, Kaiming He, Philipp
  Kr{\"{a}}henb{\"{u}}hl, and Ross~B. Girshick.
\newblock Long-term feature banks for detailed video understanding.
\newblock In {\em IEEE/CVF Conference on Computer Vision and Pattern
  Recognition}, 2019.

\bibitem{wu2022memvit}
Chao-Yuan Wu, Yanghao Li, Karttikeya Mangalam, Haoqi Fan, Bo Xiong, Jitendra
  Malik, and Christoph Feichtenhofer.
\newblock Memvit: Memory-augmented multiscale vision transformer for efficient
  long-term video recognition.
\newblock In {\em IEEE/CVF Conference on Computer Vision and Pattern
  Recognition}, 2022.

\bibitem{wu2020context}
Jianchao Wu, Zhanghui Kuang, Limin Wang, Wayne Zhang, and Gangshan Wu.
\newblock Context-aware rcnn: A baseline for action detection in videos.
\newblock In {\em European Conference on Computer Vision}, 2020.

\bibitem{wu2023stmixer}
Tao Wu, Mengqi Cao, Ziteng Gao, Gangshan Wu, and Limin Wang.
\newblock Stmixer: A one-stage sparse action detector.
\newblock In {\em IEEE/CVF Conference on Computer Vision and Pattern
  Recognition}, 2023.

\bibitem{wu2019adaframe}
Zuxuan Wu, Caiming Xiong, Chih-Yao Ma, Richard Socher, and Larry~S Davis.
\newblock Adaframe: Adaptive frame selection for fast video recognition.
\newblock In {\em IEEE/CVF Conference on Computer Vision and Pattern
  Recognition}, 2019.

\bibitem{xiao2022hierarchical}
Fanyi Xiao, Kaustav Kundu, Joseph Tighe, and Davide Modolo.
\newblock Hierarchical self-supervised representation learning for movie
  understanding.
\newblock In {\em IEEE/CVF Conference on Computer Vision and Pattern
  Recognition}, 2022.

\bibitem{DBLP:conf/cvpr/ZhangTHS19}
Yubo Zhang, Pavel Tokmakov, Martial Hebert, and Cordelia Schmid.
\newblock A structured model for action detection.
\newblock In {\em IEEE/CVF Conference on Computer Vision and Pattern
  Recognition}, 2019.

\bibitem{zhang2012slow}
Zhang Zhang and Dacheng Tao.
\newblock Slow feature analysis for human action recognition.
\newblock {\em IEEE transactions on pattern analysis and machine intelligence},
  2012.

\bibitem{zhao2022tuber}
Jiaojiao Zhao, Yanyi Zhang, Xinyu Li, Hao Chen, Bing Shuai, Mingze Xu, Chunhui
  Liu, Kaustav Kundu, Yuanjun Xiong, Davide Modolo, et~al.
\newblock Tuber: Tubelet transformer for video action detection.
\newblock In {\em IEEE/CVF Conference on Computer Vision and Pattern
  Recognition}, 2022.

\bibitem{DSN}
Yin{-}Dong Zheng, Zhaoyang Liu, Tong Lu, and Limin Wang.
\newblock Dynamic sampling networks for efficient action recognition in videos.
\newblock {\em IEEE Transactions on Image Processing}, 29:7970--7983, 2020.

\bibitem{zhi2021mgsampler}
Yuan Zhi, Zhan Tong, Limin Wang, and Gangshan Wu.
\newblock Mgsampler: An explainable sampling strategy for video action
  recognition.
\newblock In {\em IEEE/CVF International Conference on Computer Vision}, 2021.

\end{thebibliography}
}

\clearpage
\appendix
\section*{Appendix}
In this supplementary material, we provide more details of EVAD from the following aspects:
\begin{itemize}
    \item The detailed architecture illustration is in \S~\ref{sec:arch}. 
    \item The implementation details are in \S~\ref{sec:details}.
    \item Additional experimental results are in \S~\ref{sec:results}.
    \item Results analysis and visualization are in \S~\ref{sec:res}.
\end{itemize}

\section{Architectures}
\label{sec:arch}
We present the architectural details of EVAD based on 16-frame vanilla ViT-Base and ViT-Large backbones used in the experiments and the corresponding output sizes of each stage, as shown in Table~\ref{table:arch}.
The {\it token pruning} with keep rate $\rho$ is executed three times in total, following each stage. The intermediate keyframe features from backbone stages 1-4 are used for actor localization via EVAD localization branch, and the spatiotemporal features from stage 4 are used for actor feature refinement and final action classification prediction via EVAD classification branch. The computational costs of both models are reduced by 40\% (ViT-B) and 42\% (ViT-L) at $\rho$=0.7, respectively.

\setlength{\tabcolsep}{2pt}
\begin{table*}
\small
% \footnotesize
\begin{center}
\begin{tabular}{c|c|c|c} 
    \shline \textbf{Stage} & \textbf{Vision Transformer (Base)} & \textbf{Vision Transformer (Large)} & \textbf{Output Sizes}  \\
    \hline 
    data & \multicolumn{2}{c|}{stride \tcolor{4}\x\xycolor{1}\x\xycolor{1}} & \wcolor{3}\x\tcolor{16}\x\xycolor{224}\x\xycolor{224}  \\
    \hline
    \multirow{2}{*}{cube} & \tcolor{2}\x\xycolor{16}\x\xycolor{16}, {\wcolor{768} } & \tcolor{2}\x\xycolor{16}\x\xycolor{16}, {\wcolor{1024} } & \multirow{2}{*}{ \wcolor{C}\x\tcolor{8}\x\xycolor{14}\x\xycolor{14}} \\
	& stride \tcolor{2}\x\xycolor{16}\x\xycolor{16} & stride \tcolor{2}\x\xycolor{16}\x\xycolor{16} &  \\
	\hline 
    \multirow{2}{*}{stage1} & \blockatt{768}{3072}{3} & \blockatt{1024}{4096}{6} & \multirow{2}{*}{ \wcolor{C}\x[\tcolor{8}\x\xycolor{14}\x\xycolor{14}]}\\
            & &  \\
    \hline
    \multirow{3}{*}{token pruning} & \blocktoken{768}{3072}{1} & \blocktoken{1024}{4096}{1} & \multirow{3}{*}{ \wcolor{C}\x[\tcolor{8}\x\xycolor{14}\x\xycolor{14}\x\maskcolor{$\rho$}]}\\
            & &  \\
            & &  \\
    \hline
    \multirow{2}{*}{stage2} & \blockatt{768}{3072}{2} & \blockatt{1024}{4096}{5} & \multirow{2}{*}{ \wcolor{C}\x[\tcolor{8}\x\xycolor{14}\x\xycolor{14}\x\maskcolor{$\rho$}]}\\
            & &  \\
    \hline
    \multirow{3}{*}{token pruning} & \blocktoken{768}{3072}{1} & \blocktoken{1024}{4096}{1} & \multirow{3}{*}{ \wcolor{C}\x[\tcolor{8}\x\xycolor{14}\x\xycolor{14}\x\maskcolor{$\rho^{2}$}]}\\
            & &  \\
            & &  \\
    \hline
    \multirow{2}{*}{stage3} & \blockatt{768}{3072}{2} & \blockatt{1024}{4096}{5} & \multirow{2}{*}{ \wcolor{C}\x[\tcolor{8}\x\xycolor{14}\x\xycolor{14}\x\maskcolor{$\rho^{2}$}]}\\
            & &  \\
    \hline
    \multirow{3}{*}{token pruning} & \blocktoken{768}{3072}{1} & \blocktoken{1024}{4096}{1} & \multirow{3}{*}{ \wcolor{C}\x[\tcolor{8}\x\xycolor{14}\x\xycolor{14}\x\maskcolor{$\rho^{3}$}]}\\
            & &  \\
            & &  \\
    \hline
    \multirow{2}{*}{stage4} & \blockatt{768}{3072}{2} & \blockatt{1024}{4096}{5} & \multirow{2}{*}{ \wcolor{C}\x[\tcolor{8}\x\xycolor{14}\x\xycolor{14}\x\maskcolor{$\rho^{3}$}]}\\
            & &  \\
    \hline
    norm & LayerNorm(\wcolor{768}) & LayerNorm(\wcolor{1024}) & \wcolor{C}\x[\tcolor{8}\x\xycolor{14}\x\xycolor{14}\x\maskcolor{$\rho^{3}$}] \\
    \hline
    GFLOPs, \maskcolor{$\rho$}=0.7/1.0 & $134.2$ / $223.8$ & $409.4$ / $707.9$ & - \\
\shline
\end{tabular}
\caption{\textbf{Architecture details of EVAD backbone.} The {\it token pruning} denotes a transformer layer with keyframe-centric token pruning and is the same as the blocks in each stage when keep rate \maskcolor{$\rho$}=1.0. The output sizes are denoted by $\{ $$\wcolor{C}$\x$\tcolor{T}$\x$\xycolor{S}$\x$\maskcolor{\rho}$$\}$ for channel, temporal, spatial and keep rate sizes.}
\label{table:arch}
\end{center}
\vspace{-5mm}
\end{table*}
\setlength{\tabcolsep}{1.4pt}

\section{Implementation Details}
\label{sec:details}
{\flushleft \bf Query-based actor localization head.} The localization head initializes $n$ learnable proposal boxes and corresponding proposal features, which can be optimized together in the network. Then, the head utilizes RoIAlign operations to extract RoI features for each box. Next, a sequentially-stacked Dynamic Instance Interactive head~\cite{sun2021sparse,chen2021watch} is conducted on each RoI feature to generate the final predictions conditioned on proposal features. Finally, two task-specific prediction layers are used to produce the prediction boxes and corresponding actor confidence scores.

{\flushleft \bf Configurations.}
By default, the token pruning module is incorporated into the $4^{th}$, $7^{th}$, and $10^{th}$ layer of ViT-B (with 12 layers in total) and incorporated into the $7^{th}$, $13^{th}$, and $19^{th}$ layer of ViT-L (with 24 layers in total).
We specify the number of query $n$ as 100 and the dimension of query as 256, and we use 6 dynamic instance interactive modules in actor localization branch, same as in~\cite{chen2021watch}.
For action classification branch, the dimension of context refinement decoder is 384 (ViT-B) and 512 (ViT-L), and the depth of decoders for these two backbones is 6 and 12. 
The backbone is initialized with Kinetics-pretrained weights from {\it VideoMAE} and other newly added layers are initialized with Xavier~\cite{glorot2010understanding}.

{\flushleft \bf Losses and optimizers.}
For all experiments, we simply follow those in the original paper of WOO. Specifically, the loss function includes the set prediction loss and the action classification loss, where the set prediction loss consists of the cross-entropy loss over two classes (person and background), L1 loss and GIoU loss on the box. The action classification loss is denoted by the binary cross-entropy loss. We set the loss weight as $\lambda_{ce}$=2, $\lambda_{L1}$=5, $\lambda_{GIoU}$=2, $\lambda_{bce}$=12.
We use AdamW~\cite{loshchilov2017decoupled} with weight decay $1\times 10^{-4}$ as the optimizer and apply intermediate supervision on the output of each layer in localization and classification branches.

{\flushleft \bf Training and inference recipes.}
Following the $1\times$ training schedule in~\cite{sun2021sparse,chen2021watch}, we train all models for 12 epochs with an initial learning rate of $2.5\times 10^{-5}$ and reduce the learning rate by $10\times$ at epoch 5 and 8. We apply a linear warm-up from $2.5\times 10^{-6}$ to $2.5\times 10^{-5}$ at the first two epochs.
The mini-batch consists of 16 video clips and all models are trained with 8 GPUs (2 clips per device), and for the model with ViT-L backbone, both the mini-batch and the learning rate are reduced by 1/2 of the original.
For ablation studies on AVA, we set the shortest side of each video frame to 224 for efficient exploration, and for comparisons to the state-of-the-art methods, we set the shortest side to 288 unless otherwise specified. For the experiments on JHMDB, we perform random scaling to each video frame input and set its shortest side to range from 256 to 320 pixels. For the experiments on UCF101-24, we set the shortest side of each video frame to 224 for training and 256 for inference.
We perform the same training recipe as in the baseline for EVAD models under different keep rates without additional modifications, which indicates that our method can be simply incorporated into existing models and work well.

For inference, given an input video clip, EVAD directly predicts 100 proposal boxes and the corresponding person detection and action classification scores. The prediction boxes with a detection score larger than 0.7 are taken as the final results.

\section{Additional Results}
\label{sec:results}

\setlength{\tabcolsep}{2pt}
\begin{table}
\small
\begin{center}
\begin{tabular}{lccx{60}cc} 
\shline model & e2e & $T \times \tau$ & backbone & pre-train & mAP \\
\hline 
AVA~\cite{gu2018ava}$^\ast$ & \XSolidBrush & $40\times 1$ & I3D-VGG & K400 & $15.8$ \\
LFB~\cite{DBLP:conf/cvpr/WuF0HKG19} & \XSolidBrush & $32\times 2$ & I3D-R101-NL & K400 & $27.7$ \\
CA-RCNN~\cite{wu2020context} & \XSolidBrush & $32\times 2$ &  R50-NL & K400 & $28.0$ \\
SlowFast~\cite{DBLP:conf/iccv/Feichtenhofer0M19} & \XSolidBrush & $32\times 2$ & SF-R101-NL & K600 & $28.2$ \\
ACAR-Net~\cite{pan2021actor} & \XSolidBrush & $32\times 2$ & SF-R101-NL & K400 & $30.0$ \\
AIA~\cite{DBLP:conf/eccv/TangXMPL20} & \XSolidBrush & $32\times 2$ & SF-R101 & K700 & $31.2$ \\
ACRN~\cite{sun2018actor}$^\ast$ & \Checkmark & $20\times 1$ & S3D-G & K400 & $17.4$ \\
VAT~\cite{girdhar2019video} & \Checkmark & $64\times 1$ & I3D-VGG & K400 & $25.0$ \\
WOO~\cite{chen2021watch} & \Checkmark & $32\times 2$ & SF-R101-NL & K600 & $28.0$ \\
TubeR~\cite{zhao2022tuber} & \Checkmark & $32\times 2$ & CSN-152 & IG+K400 & $31.7$ \\
STMixer~\cite{wu2023stmixer} & \Checkmark & $32\times 2$ & CSN-152 & IG+K400 & $34.4$ \\
\hline $\mathbf{EVAD}$, $\rho$=0.7 & \Checkmark & $16\times 4$ & ViT-B & K400 & $31.1$ \\
$\mathbf{EVAD}$, $\rho$=0.7 & \Checkmark & $16\times 4$ & ViT-L & K700 & $\mathbf{38.7}$  \\
\shline
\end{tabular}
\caption{\textbf{Comparison with the state-of-the-art on AVA v2.1.} \Checkmark denotes an end-to-end approach using a unified backbone, and \XSolidBrush denotes a two-stage approach using two separated backbones. $T \times \tau$ refers to the frame number and corresponding sample rate. Methods marked with $^\ast$ leverage optical flow input.}
\label{table:ava1_sota}
\end{center}
\vspace{-7mm}
\end{table}
\setlength{\tabcolsep}{1.4pt}

We compare our EVAD with the state-of-the-art methods on AVA v2.1 in Table~\ref{table:ava1_sota}. With fewer input frames, EVAD with ViT-B backbone outperforms most two-stage and end-to-end models and has comparable performance to AIA with mAP of $31.1$ vs. $31.2$. When we apply a larger backbone ViT-L and use the same pre-trained dataset as AIA, the performance can surpass AIA by a large margin. Also, it outperforms the newly end-to-end methods TubeR and STMixer, the latter two equipped with long-term feature banks.

\setlength{\tabcolsep}{2pt}
\begin{table}
\small
  \begin{center}
      \begin{tabular}{c|cccc|cc}
      \shline
      \multirow{2}{*}{model} &\multicolumn{4}{c|}{UCF24} & \multicolumn{2}{c}{JHMDB }  \\
      \cline{2-7}
      & f-mAP & 0.20 & 0.50 & 0.50:0.95 & 0.20 & 0.50\\
      \hline
      WOO$^\ddag$~\cite{chen2021watch} & 76.7 & 74.4 & 55.8 & 26.0 & 70.0 & 69.5\\
      TubeR$^\ast$~\cite{zhao2022tuber} (I3D) & 81.3 & \bf{85.3} & \bf{60.2} & 29.7 & 81.8 & 80.7\\
      TubeR~\cite{zhao2022tuber} (CSN-152) & 83.2 & 83.3 & 58.4 & 28.9 & \bf{87.4} & \bf{82.3}\\
      \hline EVAD, $\rho$=1.0 & 84.9 & 76.6 & 60.1 & \bf{30.0} & 78.2 & 77.1\\
      EVAD, $\rho$=0.6 & \bf{85.1} & 76.4 & 58.8 & 29.1 & 79.0 & 77.8\\
      \shline
      \end{tabular}
    \caption{\textbf{Comparison on UCF24 and JHMDB with video-mAP.} $^\ddag$ indicates our implementation. Methods marked with $^\ast$ leverage optical flow input.}
    \vspace{-7mm}
    \label{table:vmap}
  \end{center}
\end{table}
\setlength{\tabcolsep}{1.4pt}

Next, we provide the video-mAP results on UCF101-24 and JHMDB. As shown in Table~\ref{table:vmap}, we compare our EVAD with the state-of-the-art frame-level detector WOO and tubelet-level detector TubeR. Our method follows the pipeline of WOO and is also a frame-level detector. We achieve better performance than WOO under various settings of the video-mAP. Without using tubelet annotations for training, our method has the lower performance than TubeR at multiple IoU thresholds. However, we observe that our EVAD achieves on-par or even better performance when increasing IoU thresholds on UCF101-24. This indicates that our EVAD can generate high-quality action tubes.

\begin{figure*}[ht!]
\centering
\includegraphics[width=1.0\linewidth]{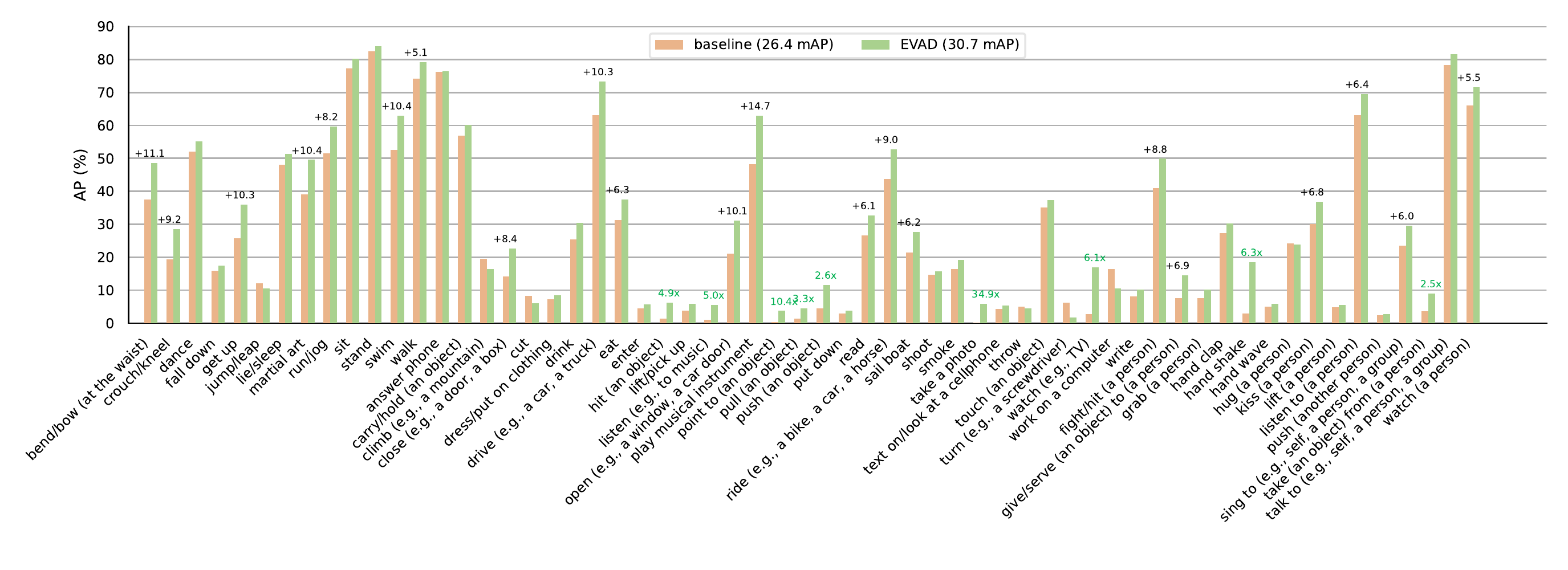}
\vspace{-6mm}
\caption{\textbf{Per-category AP for ViT baseline (26.4 mAP) and EVAD (30.7 mAP) on AVA.} Categories that increase in absolute value by more than 5\% are marked in \textbf{\textcolor[RGB]{0,0,0}{black}} and those more than twice the AP of the baseline are marked in \textbf{\textcolor[RGB]{0,176,80}{green}}.}
\label{fig:per_ap}
\end{figure*}

\section{Result Analysis and Visualization}
\label{sec:res}
Firstly, we compare the per-category performance of a ViT baseline and our EVAD on AVA, as shown in Fig.~\ref{fig:per_ap}. Our method improves in 53 out of 60 categories, with significant improvements for categories with fast movement (e.g., {\it bend/bow (at the waist)} (+11.1\%) and {\it martial art} (+10.4\%)) and categories with scene interaction (e.g., {\it drive (e.g., a car, a truck)} (+10.3\%) and {\it play musical instrument} (+14.7\%)).
This illustrates the effectiveness of two core designs of our EVAD, i.e., the proposed keyframe-centric token pruning can preserve tokens that contain sufficient action semantics, and these preserved tokens can enrich each actor spatiotemporal and scene feature through the proposed context refinement decoder.

\begin{figure*}[t]
\centering
		\begin{center}
		\subfloat[Categories that EVAD with token pruning outperforms EVAD without token pruning by more than 0.5\%.]{
			\begin{minipage}{15.5cm}
				\includegraphics[width=15.5cm]{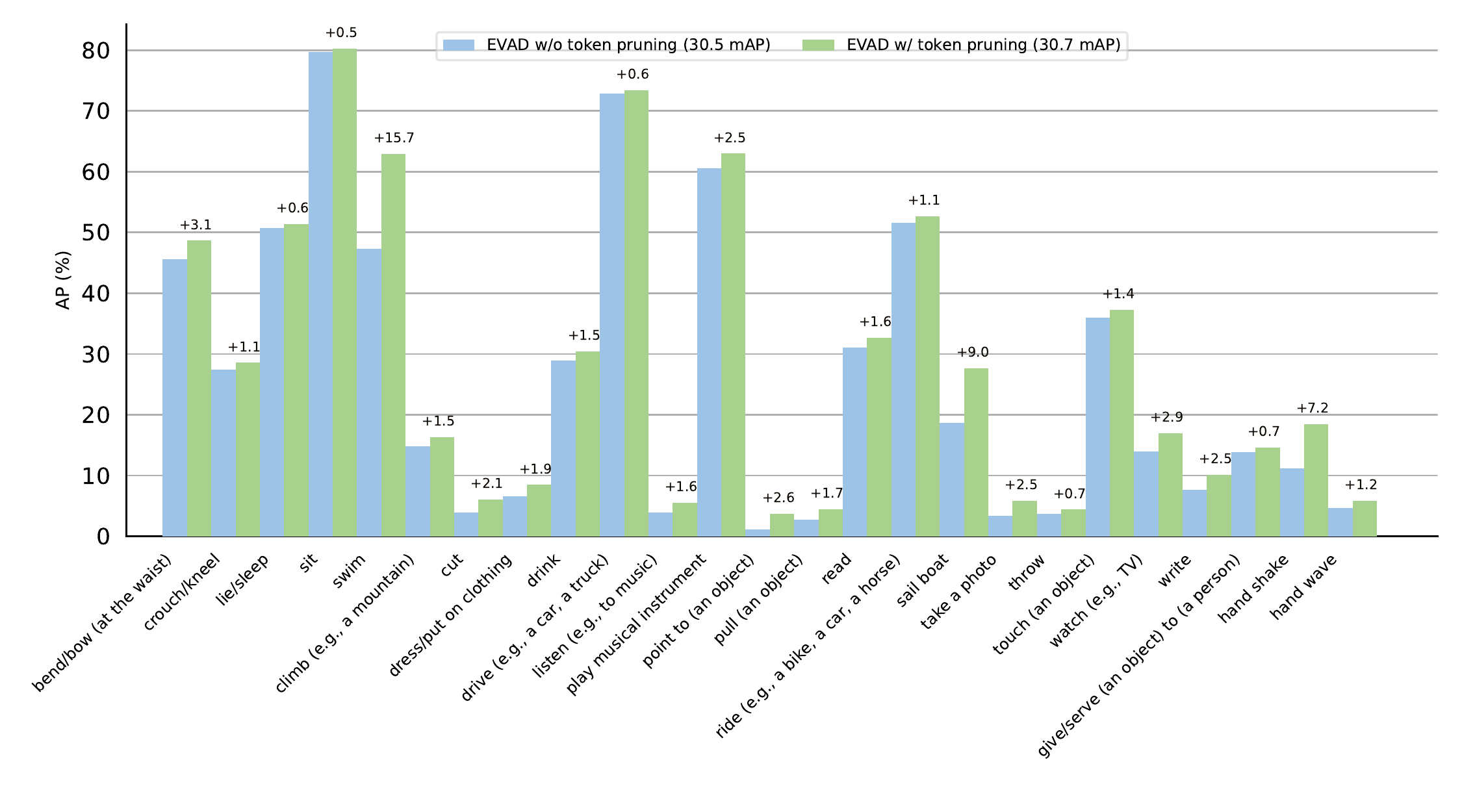}
			\end{minipage}
			\label{fig:better}
		}
        \quad
		\subfloat[Categories that EVAD without token pruning performs better.]{
			\begin{minipage}{15.5cm}
				\includegraphics[width=15.5cm]{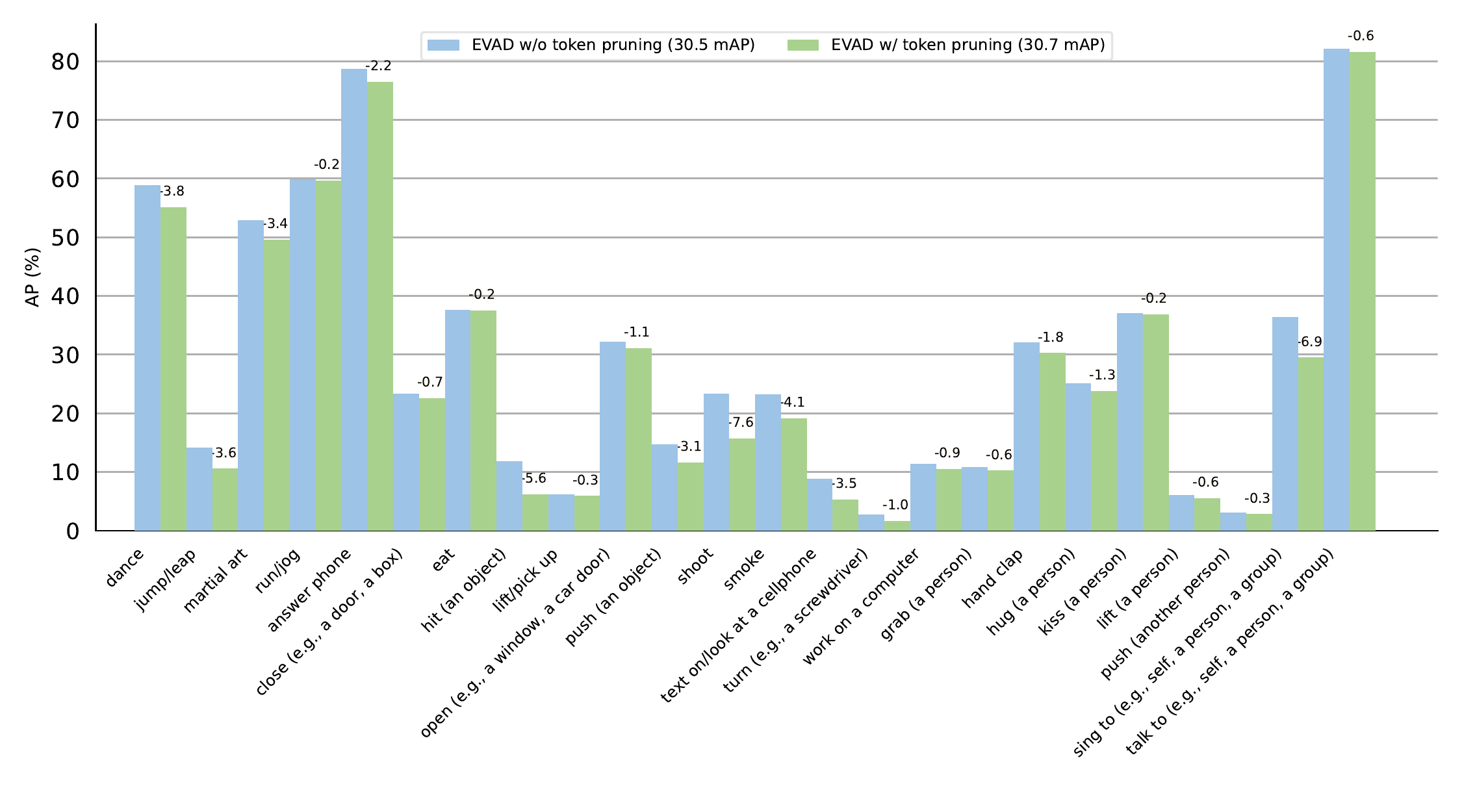}
			\end{minipage}
			\label{fig:worse}
		}
		\end{center}
		\vspace{-5mm}
		   \caption{\textbf{Per-category AP for EVAD w/o token pruning (30.5mAP) and EVAD w/ token pruning (30.7 mAP) on AVA.}}
            \label{fig:compare}
            \vspace{-3mm}
\end{figure*}

As seen in the experimental results, our keyframe-centric token pruning enables EVAD to achieve comparable performance to its counterpart without pruning, and we further compare the performance of these two models on each category of AVA, as shown in Fig.~\ref{fig:compare}.
We observe that although the overall performance of two models is comparable, the performance trend of them is inconsistent on each category. Concretely, EVAD with token pruning performance increases a lot in {\it swim} (+15.7\%), {\it sail boat} (+9.0\%), and {\it hand shake} (+7.2\%) categories, and decreases a lot in {\it hit} (-5.6\%), {\it shoot} (-7.6\%), and {\it sing to} (-6.9\%).
We consider that token pruning drops a high percentage of tokens (66\%), resulting in poor performance on categories with small motion or interaction with small objects, and good performance on categories with opposite characteristics.

\begin{figure*}[t]
\centering
\vspace{-3mm}
\includegraphics[width=0.88\linewidth]{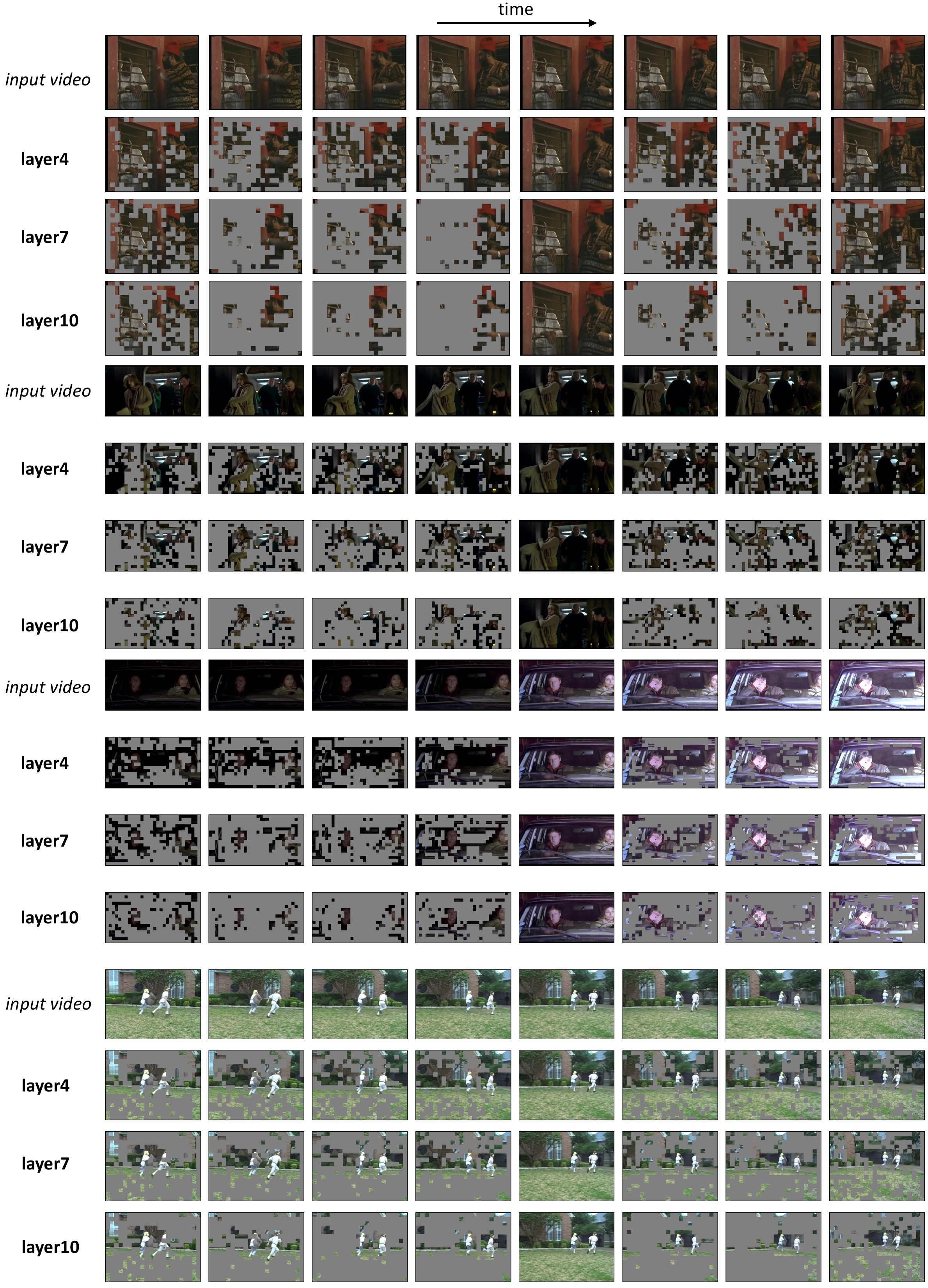}
\vspace{-1mm}
\caption{\textbf{More visualization of preserved tokens by encoder layers with keyframe-centric token pruning on AVA.}  Given an input of 8-frame intermediate tokens, the keyframe tokens are all retained on the fifth column, and redundant tokens in other frames are progressively removed.}
\label{fig:enc_vis}
\vspace{-3mm}
\end{figure*}

To show the effectiveness of our token pruning method for retaining semantic cues, we collect more visualizations of token pruning as a supplementary of Figure 4 in our paper, as shown in Fig.~\ref{fig:enc_vis}. EVAD is able to preserve important tokens in non-keyframes, e.g., for the person putting on clothing in {\it example 2}, it can preserve the sleeve with a large movement deformation.
Moreover, we observe that those frames further away from the keyframe retain a greater number of tokens in most examples. Due to the slowness of video semantics varying in the temporal dimension~\cite{zhang2012slow}, frames adjacent to the keyframe have higher semantic redundancy. When we perform keyframe-centric tokens pruning, more tokens from adjacent frames are discarded.

\begin{figure*}[t]
\centering
\vspace{-3mm}
\includegraphics[width=0.9\linewidth]{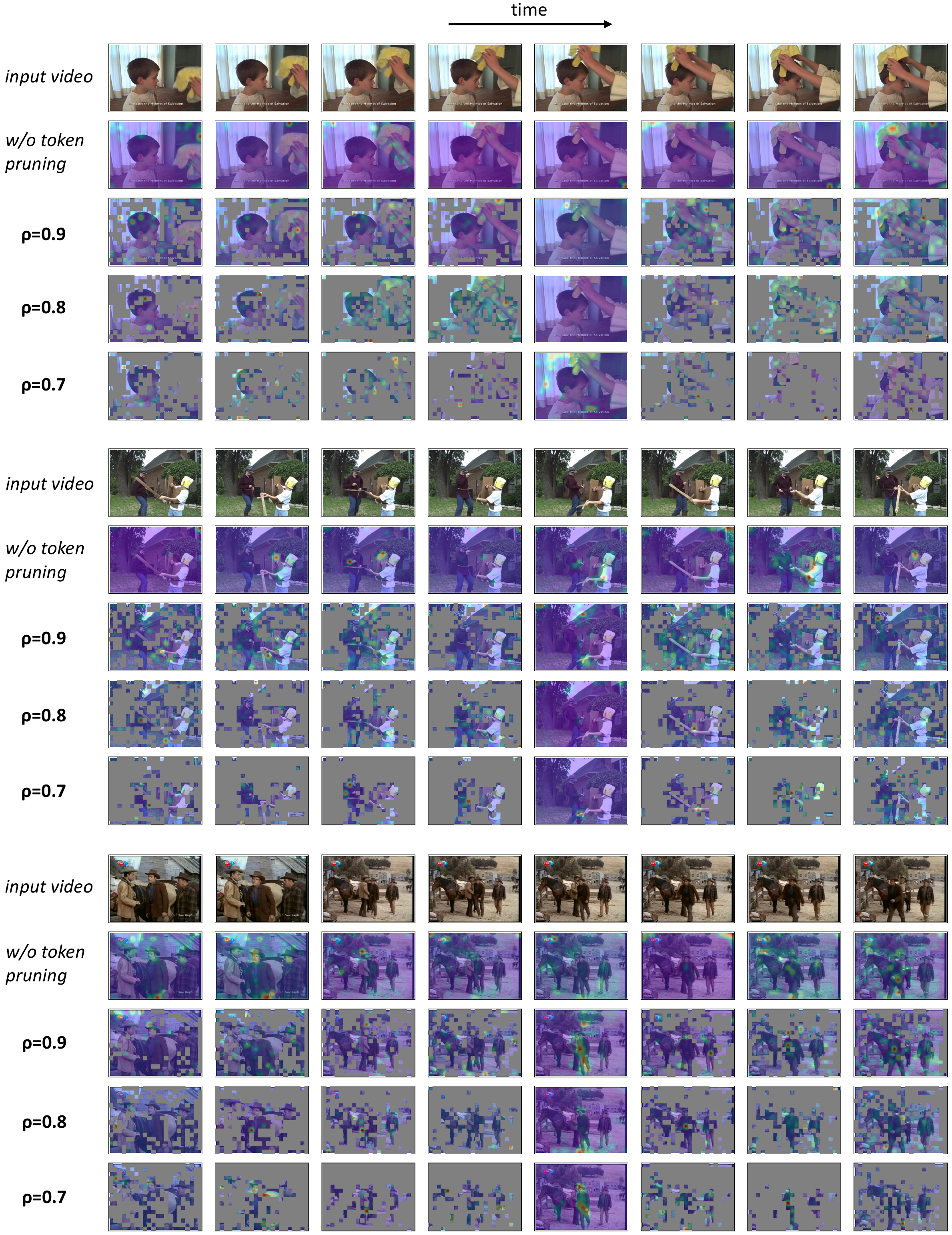}
\vspace{-1mm}
\caption{\textbf{Visualization of context refinement decoder attention maps on AVA.} For each example, 1st row: RGB frames of the input clip with a stride of 8 for viewing better, where the fifth image represents the keyframe. 2nd row: attention maps of the last decoder layer without token pruning between actors of interest and the whole video tokens.  3rd to 5th rows: attention maps of the last layer at different token keep rates ($\rho$=0.9/0.8/0.7) between actors of interest and the preserved tokens.}
\label{fig:dec_vis}
\vspace{-3mm}
\end{figure*}

Finally, to illustrate that using preserved video tokens for context refinement can maintain the same performance as using the whole video tokens, we visualize the attention maps of our context refinement decoder at different token keep rates, as shown in Fig.~\ref{fig:dec_vis}, where the attention result is the average of the attention maps between $n$ actors of interest and $M$ video tokens.
We observe that those regions with high attentive values of the decoder without token pruning can be preserved at various keep rates, e.g., the hat and the wearing hand in {\it example 1}.
This further demonstrates that our token pruning can retain semantic information for action classification and the proposed context refinement decoder can enrich actor features via remaining context, which maintain the detection accuracy.

\end{document}